\documentclass[a4paper, 11pt]{article}
\pdfoutput=1  
\usepackage{etoolbox,verbatim}
\usepackage{subcaption}
\newtoggle{arxiv}
\toggletrue{arxiv} 

\newtoggle{customthms}
\toggletrue{customthms} 

\newtoggle{colt}
\togglefalse{colt} 

\usepackage{xcolor}
\usepackage{tcolorbox}
\tcbuselibrary{skins,breakable}

\tcbset{
  aibox/.style={
      width=\linewidth,
      top=6pt,
      bottom=0pt,
      left=1.5pt,
      right=1.5pt,
      colback=blue!60!gray!5,
      colframe=black,
      colbacktitle=black,
      enhanced,
      center,
      attach boxed title to top left={yshift=-0.1in,xshift=0.15in},
      boxed title style={boxrule=0pt,colframe=white,},
    }
}
\newtcolorbox{AIbox}[2][]{aibox,title=#2,#1}

\definecolor{primalcolor}{HTML}{A60000}
\definecolor{contrarycolor}{HTML}{00A6A6}
\definecolor{darkcontrarycolor}{HTML}{004C4C}
\definecolor{lightblue}{HTML}{2970CC}
\definecolor{lightpurple}{HTML}{673147}
\definecolor{ForestGreen}{HTML}{FF5733}
\definecolor{myred}{HTML}{AA4A44}
\definecolor{hyppurple}{HTML}{800080}

\newcommand{\linkcolor}{darkcontrarycolor}
\newcommand{\urlcolor}{darkcontrarycolor}
\newcommand{\citecolor}{darkcontrarycolor}

\newcommand{\thmcolordark}{red!30!black}
\newcommand{\lightred}{red!40!white}

\usepackage[T1]{fontenc}
\usepackage{capt-of}
\usepackage[font=small,labelfont=bf]{caption}

\usepackage{natbib}
\usepackage{scalefnt}
\usepackage{amssymb,upgreek,bm}
\usepackage{mathtools}
\usepackage[english]{babel}  
\usepackage{multirow,tcolorbox,tikz-cd,turnstile,xspace,xparse}
\usepackage{relsize,breakcites,appendix}
\usepackage{longtable,tablefootnote,booktabs,float,makecell} 
\usepackage[normalem]{ulem}
\usepackage{tabu}
\usepackage[shortlabels]{enumitem} 
\usepackage{footnote}
\usepackage{boxedminipage}
\usepackage{multirow,nicefrac}
\usepackage{verbatim} 
\usepackage{wrapfig}

\iftoggle{arxiv}
{
\usepackage[colorlinks=true,linkcolor=\linkcolor,urlcolor=\urlcolor,citecolor=\citecolor,breaklinks]{hyperref}
}
{
\hypersetup{colorlinks=true,linkcolor=\linkcolor,urlcolor=\urlcolor,citecolor=\citecolor,breaklinks}
}

\usepackage{prettyref}

\usepackage[capitalise,nameinlink]{cleveref}

\iftoggle{colt}{
    \DeclareRobustCommand{\qed}{
        \usepackage{thmtools}
          \ifmmode \mathqed
          \else
            \leavevmode\unskip\penalty9999 \hbox{}\nobreak\hfill
            \quad\hbox{\qedsymbol}%
          \fi
    }
}
{
\usepackage{amsthm}
     
}
\usepackage{thm-restate}

\iftoggle{arxiv}
{
    \usepackage{mathrsfs}

    \usepackage[
  letterpaper,
  left=1in,
  right=1in,
  top=0.9in,
  bottom=0.9in,
  headheight=14pt,
  headsep=14pt,
  footskip=20pt
]{geometry}

\usepackage{fancyhdr}
\usepackage{lipsum} 

\pagestyle{fancy}
\fancyhf{} 
\fancyfoot[C]{\small \thepage}

\renewcommand{\headrulewidth}{0.4pt}
\renewcommand{\footrulewidth}{0pt}

    \usepackage[bitstream-charter]{mathdesign}
     
    \usepackage[scaled=0.92]{PTSans}

    \usepackage[noend]{algpseudocode}
    \usepackage{algorithm}
   
}
{
    \usepackage{mathrsfs}
}

\DeclareMathAlphabet{\mathbfsf}{\encodingdefault}{\sfdefault}{bx}{n}

\numberwithin{equation}{section}

\iftoggle{arxiv}
{
    \newcommand{\colorbold}[1]{
    \textbf{\textcolor{\thmcolor}{#1}}}}
{
    \newcommand{\colorbold}[1]{\textbf{#1}
}
}
\iftoggle{arxiv}
{
    \newcommand{\colorpar}[1]{
    \paragraph{\textcolor{\thmcolor}{#1}}}
}
{
    \newcommand{\colorpar}[1]{\paragraph{#1}}
}

\usepackage{thmtools}

\Crefname{equation}{Eq.}{Eqs.}
\Crefname{assumption}{Assumption}{Assumptions}
\Crefname{condition}{Condition}{Conditions}
\Crefname{claim}{Claim}{Claims}
\Crefname{property}{Property}{Properties}
\Crefname{construction}{Construction}{Constructions}

\declaretheoremstyle[
    headformat=\normalfont\textcolor{\thmcolordark}{\bfseries\NAME\,\NUMBER}\NOTE,%
    notefont={\normalfont\textcolor{\thmcolordark}{\bfseries}}, 
    notebraces={}{},
    bodyfont=\normalfont\itshape,
    spaceabove = 6pt,
    spacebelow = 6pt,
    ]{coloredthmversion}

\declaretheoremstyle[
    headformat=\normalfont\textcolor{\thmcolordark}{\bfseries\NAME\,\NUMBER}\NOTE,%
    bodyfont=\normalfont\itshape,
    spaceabove = 6pt,
    spacebelow = 6pt,
    ]{coloredthm}

\declaretheoremstyle[
    headformat=\normalfont\textcolor{\thmcolordark}{\bfseries\NAME\,\NUMBER}\NOTE,%
    bodyfont=\normalfont,
    spaceabove = 6pt,
    spacebelow = 6pt,
    ]{coloreddef}

\iftoggle{customthms}
{
    \theoremstyle{coloredthmversion}
}
{}

\iftoggle{customthms}{
  \theoremstyle{coloredthm}
  \newtheorem{theorem}{Theorem}
  \newtheorem{lemma}{Lemma}[section]
  \newtheorem{corollary}{Corollary}[section]
  \newtheorem{proposition}[lemma]{Proposition}
}
{}

\newtheorem*{thminformal*}{Informal Theorem}

\iftoggle{customthms}{
    \theoremstyle{coloreddef}
    \newtheorem{definition}{Definition}[section]

    \newtheorem{property}{Property}[section]
}
{
  
}

\newtheorem{assumption}{Assumption}[section]
\newtheorem{condition}{Condition}[section]

\makeatletter
\newcommand{\neutralize}[1]{\expandafter\let\csname c@#1\endcsname\count@}
\makeatother

\iftoggle{customthms}{
    \newtheoremstyle{named}{}{}{\itshape}{}{\bfseries}{}{.5em}{\Cref{#3} {\normalfont (informal)} }{}
    \theoremstyle{named}
    
    \theoremstyle{plain}
}
{}

\newtheorem*{theorem*}{Theorem}
\newtheorem*{lemma*}{Lemma}
\newtheorem*{corollary*}{Corollary}
\newtheorem*{proposition*}{Proposition}
\newtheorem*{claim*}{Claim}
\newtheorem*{fact*}{Fact}
\newtheorem*{observation*}{Observation}
\newtheorem*{definition*}{Definition}
\newtheorem*{remark*}{Remark}
\newtheorem*{example*}{Example}

\def\ddefloop#1{\ifx\ddefloop#1\else\ddef{#1}\expandafter\ddefloop\fi}
\def\ddef#1{\expandafter\def\csname bb#1\endcsname{\ensuremath{\mathbb{#1}}}}
\ddefloop ABCDEFGHIJKLMNOPQRSTUVWXYZ\ddefloop

\def\ddefloop#1{\ifx\ddefloop#1\else\ddef{#1}\expandafter\ddefloop\fi}
\def\ddef#1{\expandafter\def\csname frak#1\endcsname{\ensuremath{\mathfrak{#1}}}}
\ddefloop ABCDEFGHIJKLMNOPQRSTUVWXYZ\ddefloop

\def\ddefloop#1{\ifx\ddefloop#1\else\ddef{#1}\expandafter\ddefloop\fi}
\def\ddef#1{\expandafter\def\csname fr#1\endcsname{\ensuremath{\mathfrak{#1}}}}
\ddefloop ABCDEFGHIJKLMNOPQRSTUVWXYZ\ddefloop

\def\ddefloop#1{\ifx\ddefloop#1\else\ddef{#1}\expandafter\ddefloop\fi}
\def\ddef#1{\expandafter\def\csname eul#1\endcsname{\ensuremath{\EuScript{#1}}}}
\ddefloop ABCDEFGHIJKLMNOPQRSTUVWXYZ\ddefloop

\def\ddefloop#1{\ifx\ddefloop#1\else\ddef{#1}\expandafter\ddefloop\fi}
\def\ddef#1{\expandafter\def\csname scr#1\endcsname{\ensuremath{\mathscr{#1}}}}
\ddefloop ABCDEFGHIJKLMNOPQRSTUVWXYZ\ddefloop

\def\ddefloop#1{\ifx\ddefloop#1\else\ddef{#1}\expandafter\ddefloop\fi}
\def\ddef#1{\expandafter\def\csname b#1\endcsname{\ensuremath{\mathbf{#1}}}}
\ddefloop ABCDEFGHIJKLMNOPQRSTUVWXYZ\ddefloop

\def\ddefloop#1{\ifx\ddefloop#1\else\ddef{#1}\expandafter\ddefloop\fi}
\def\ddef#1{\expandafter\def\csname bhat#1\endcsname{\ensuremath{\hat{\mathbf{#1}}}}}
\ddefloop ABCDEFGHIJKLMNOPQRSTUVWXYZ\ddefloop

\def\ddefloop#1{\ifx\ddefloop#1\else\ddef{#1}\expandafter\ddefloop\fi}
\def\ddef#1{\expandafter\def\csname btil#1\endcsname{\ensuremath{\tilde{\mathbf{#1}}}}}
\ddefloop ABCDEFGHIJKLMNOPQRSTUVWXYZ\ddefloop

\def\ddefloop#1{\ifx\ddefloop#1\else\ddef{#1}\expandafter\ddefloop\fi}
\def\ddef#1{\expandafter\def\csname bst#1\endcsname{\ensuremath{\mathbf{#1}^\star}}}
\ddefloop ABCDEFGHIJKLMNOPQRSTUVWXYZ\ddefloop

\def\ddefloop#1{\ifx\ddefloop#1\else\ddef{#1}\expandafter\ddefloop\fi}
\def\ddef#1{\expandafter\def\csname bst#1\endcsname{\ensuremath{\mathbf{#1}^\star}}}
\ddefloop abcdeghijklmnopqrstuvwxyz\ddefloop

\def\ddefloop#1{\ifx\ddefloop#1\else\ddef{#1}\expandafter\ddefloop\fi}
\def\ddef#1{\expandafter\def\csname bhat#1\endcsname{\ensuremath{\hat{\mathbf{#1}}}}}
\ddefloop abcdefghijklmnopqrstuvwxyz\ddefloop

\def\ddefloop#1{\ifx\ddefloop#1\else\ddef{#1}\expandafter\ddefloop\fi}
\def\ddef#1{\expandafter\def\csname b#1\endcsname{\ensuremath{\mathbf{#1}}}}
\ddefloop abcdeghijklnopqrstuvwxyz\ddefloop

\def\ddefloop#1{\ifx\ddefloop#1\else\ddef{#1}\expandafter\ddefloop\fi}
\def\ddef#1{\expandafter\def\csname barb#1\endcsname{\ensuremath{\bar{\mathbf{#1}}}}}
\ddefloop abcdefghijklmnopqrstuvwxyz\ddefloop

\def\ddef#1{\expandafter\def\csname c#1\endcsname{\ensuremath{\mathcal{#1}}}}
\ddefloop ABCDEFGHIJKLMNOPQRSTUVWXYZ\ddefloop
\def\ddef#1{\expandafter\def\csname h#1\endcsname{\ensuremath{\widehat{#1}}}}
\ddefloop ABCDEFGHIJKLMNOPQRSTUVWXYZ\ddefloop
\def\ddef#1{\expandafter\def\csname hc#1\endcsname{\ensuremath{\widehat{\mathcal{#1}}}}}
\ddefloop ABCDEFGHIJKLMNOPQRSTUVWXYZ\ddefloop
\def\ddef#1{\expandafter\def\csname t#1\endcsname{\ensuremath{\widetilde{#1}}}}
\ddefloop ABCDEFGHIJKLMNOPQRSTUVWXYZ\ddefloop
\def\ddef#1{\expandafter\def\csname tc#1\endcsname{\ensuremath{\widetilde{\mathcal{#1}}}}}
\ddefloop ABCDEFGHIJKLMNOPQRSTUVWXYZ\ddefloop

\iftoggle{arxiv}
{
\newcommand{\togglepar}[1]{\colorpar{#1}}
}
{
\newcommand{\togglepar}[1]{\textbf{#1}}
}

\usepackage{tcolorbox}
\tcbuselibrary{skins,breakable}

\tcbset{
  aibox/.style={
      width=\linewidth,
      top=6pt,
      bottom=0pt,
      left=1.5pt,
      right=1.5pt,
colback=blue!60!gray!5,
colframe=black,
      colbacktitle=black,
      enhanced,
      center,
      attach boxed title to top left={yshift=-0.1in,xshift=0.15in},
      boxed title style={boxrule=0pt,colframe=white,},
    }
}

\newcommand{\I}{\mathbf{I}}

\newcommand{\Normal}{\mathrm{N}}

\newcommand{\eye}{\mathbf{I}}

\newcommand{\ballkr}[1][r]{\cB_{k}(r)}

\DeclareMathSymbol{\shortminus}{\mathbin}{AMSa}{"39}

\newcommand{\cubedrop}{\textsc{CubeDrop}}

\newcommand{\drawer}
{\textsc{DrawerRecall}}
\newcommand{\balance}
{\textsc{BalanceBar}}

\newcommand{\half}
{\textsc{HalfAndHalf}}

\newcommand{\algofont}[1]{{\scalefont{1.1}{\texttt{\textbf{#1}}}}}

\newcommand{\wkspcolor}{red!70!black}
\newcommand{\wkspnc}{\algofont{Wksp}}
\newcommand{\wksp}{{\color{\wkspcolor}\wkspnc}}

\newcommand{\dpnc}{\algofont{VanillaDP}}

\newcommand{\dphistcolor}{teal!80!blue}
\newcommand{\dphistnc}{\algofont{HistoryDP}}
\newcommand{\dphist}{{\color{\dphistcolor}\dphistnc}}

\newcommand{\keyframecolor}{magenta!80!black}
\newcommand{\keyframenc}{\algofont{Keyframe}}
\newcommand{\keyframe}{{\color{\keyframecolor}\keyframenc}}

\usepackage{preamble/color-edits}

\addauthor{ms}{magenta}
\addauthor{dc}{green}

\newcommand{\ignore}[1]{}

\iftoggle{arxiv}
{
\input{preamble/arxiv_title}
}
{}

\usepackage{subcaption}
\usepackage{titlesec}
\usepackage{placeins}
\usepackage{caption}

\iftoggle{arxiv}
{}
{
\titlespacing*{\section}{0pt}{0.6em}{0.3em}
\titlespacing*{\subsection}{0pt}{0.4em}{0.2em}
\titlespacing*{\subsubsection}{0pt}{0.3em}{0.15em}

\setlength{\textfloatsep}{8pt plus 2pt minus 2pt}
\setlength{\floatsep}{6pt plus 2pt minus 2pt}
\setlength{\intextsep}{6pt plus 2pt minus 2pt}

\usepackage[most]{tcolorbox}
\usepackage{xcolor}
\captionsetup{
  skip=3pt,
  belowskip=0pt,
  aboveskip=3pt
}
}

\newtcolorbox{promptbox}[1]{
  colback=gray!5,
  colframe=gray!60,
  fonttitle=\bfseries,
  title=#1,
  boxrule=0.5pt,
  arc=2mm,
  left=1mm,
  right=1mm,
  top=1mm,
  bottom=1mm
}

\title{Workspace Models: Lightweight Robotic  Memory via Saliency-Driven Supervision}

\author{
\small
Nitish Dashora\footnote{\texttt{\{dashora, idanshen, jmgola, pulkitag\}@mit.edu}\label{foot:mit}}\textsuperscript{,$\$$},~
Douglas Chen\footnote{\texttt{\{dchen3, msimchow\}@andrew.cmu.edu} \label{foot:cmu}}\textsuperscript{,$\$$},~
Idan Shenfeld\textsuperscript{\ref{foot:mit}},~
John Marangola\textsuperscript{\ref{foot:mit}},~\\ \small
Pulkit Agrawal\textsuperscript{\ref{foot:mit},$\dagger$},~
Max Simchowitz\textsuperscript{\ref{foot:cmu},$\dagger$}~\\
\vspace{-.3em}
 \rule{.38\textwidth}{.7pt}
\\
\footnotesize
$^{\$}$Project lead. $^\dagger$Equal advising. \\
\footnotesize
$^{a}$ Massachusetts Institute of Technology ~~
$^{b}$ Carnegie Mellon University ~~
}
\vspace{-.5em}
\paperdate{\today}
\date{\vspace{-.5em}}

\begin{document}

\begin{tcolorbox}[
colback=blue!60!gray!5, colframe=gray!50, 
boxrule=0pt, 
    arc=2mm%
]
\maketitle
\vspace{-1em}
\tcbline
\begin{minipage}[t]{1.0\linewidth}
    \centering
\includegraphics[width=\textwidth]{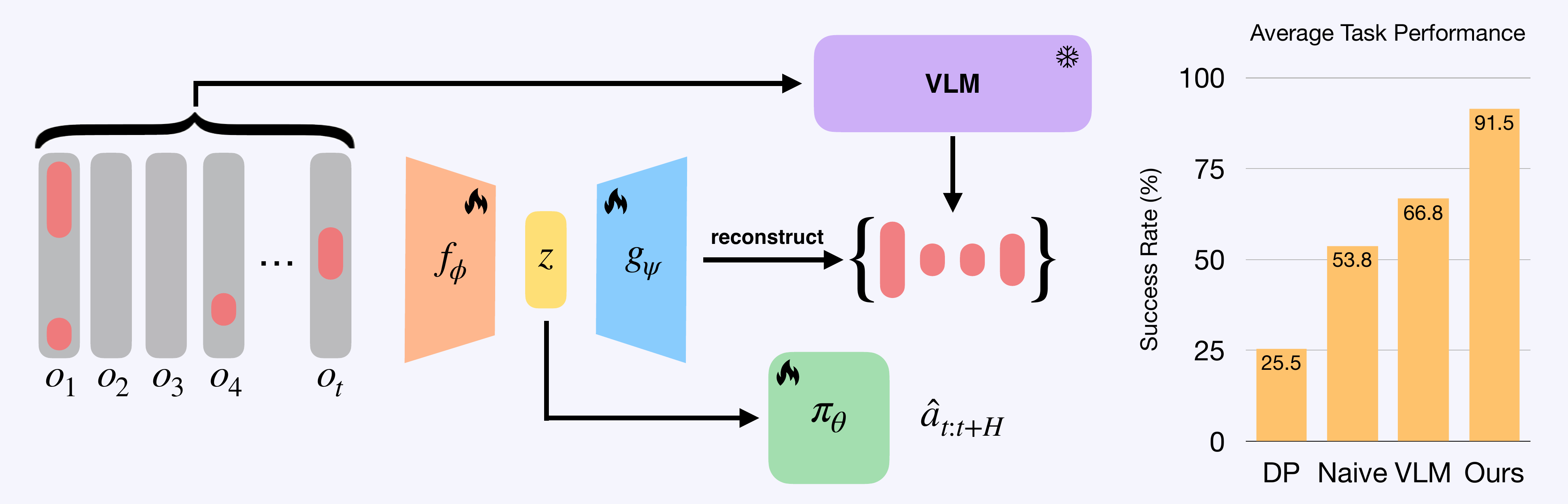}
             \captionof{figure}{
        \small  We propose the \textbf{Workspace Model}, an architecture trained to produce compressed latent representations of history used for downstream policy learning. Workspace Models are trained by encoding raw histories and reconstructing high-level salient information, as judged by a VLM, thus amortizing test-time compute. Across real and simulated tasks, our model solves memory-intensive tasks with low latency.} 
    \label{fig:teaser}
\end{minipage}

  \vskip 0.1cm
  \makeatletter
  \tcbline \ifdefempty{\metadatalist}{}{\metadatalist\par}
  \makeatother
  
\end{tcolorbox}

\begin{abstract}
    Complex robotic manipulation tasks frequently require a long-term memory of past events and actions. As conditioning on full histories renders policies prone to spurious correlations and degrades performance, many approaches to policy memory involve compressing historical information through expensive VLM queries in-the-loop to process only task-salient information. In this paper, we propose an alternative approach in which computationally intensive VLM queries are made during train-time to learn a lightweight latent memory that can be efficiently queried at deployment time. Our representation, which we call the \textbf{workspace token}, is trained by (1) using a VLM to identify current and historical information necessary for completing a task, then (2) distilling these into the workspace token using a set-reconstruction decoder loss. In both simulation and hardware, we show that the workspace token can be used as a drop-in replacement for observations during deployment, enabling policies to solve memory-intensive tasks without the need for VLM reasoning in-the-loop, in effect serving as a \textbf{latent harness} for distilling a stronger reasoning models ability to solve long-horizon tasks  to  a reactive robotic policy. We further demonstrate that the workspace tokens are not only more lightweight, but also lead to better policy performance compared to conditioning policies on explicit modalities like curated past image frames, motivating a ``latent'' approach to history curation and reasoning model harnesses more broadly.
    
\end{abstract}

\section{Introduction}

\label{sec:intro}

Solving hard decision-making problems requires memory of long temporal windows, as relevant information for control may be contained in prior observations \citep{kaelbling1998pomdps}. Despite the need for memory, today's large-scale robot architectures are conditioned on either a single \citep{black2024pi_0, pmlr-v229-zitkovich23a, kim2024openvlaopensourcevisionlanguageactionmodel,intelligence2025pi05} or a short subsequence of past observations \citep{chi2023diffusion, zhao2023learningfinegrainedbimanualmanipulation, shafiullah2022behavior}. This is because naively increasing observation history can introduce generalization error during training \citep{contextrot} as irrelevant past frames make learned behaviors susceptible to latching onto spurious correlations with current actions and can also cause the model to encounter histories outside the policy's training distribution \citep{Geirhos_2020, de2019causal}. 
{Distribution shift is worsened by compounding online error typical in imitation learning \citep{ross2010efficient, ross2011reduction,simchowitz2025pitfalls}.}
{These challenges are exacerbated in robotic applications, where data is considerably more limited and high dimensional than in natural language processing.}

The challenges of memory have motivated numerous attempts to produce {\textbf{summaries} of task history which contain what we term \textbf{salient information}: a limited amount of information from past frames needed to complete the task. Such compressed histories are still sufficient for task success, but limit  extraneous or distracting information that may harm generalization.} %
The most popular approach has been to deploy a powerful pre-trained model, such as a vision-language model (VLM), alongside the robot to provide explicit natural-language summaries of past events \citep{torne2026memmultiscaleembodiedmemory} or to select a limited number of relevant observation frames \citep{sridhar2025memerscalingmemoryrobot, mark2026bpplongcontextrobotimitation}.

However, querying powerful models in robotics comes at a cost. Reactive policies must continually make queries, incurring high computational cost and suffering from latency that slows robot policies, introduces aliasing effects, and renders deployments more brittle. Moreover, imperfect summaries, e.g. those that lack key salient information, limit policy performance. This introduces a tradeoff where larger or more powerful models can provide more reliable summaries, but introduce greater latency at test-time, leading to the issues described above. 
{For example, a medium-sized VLM \citep{gemmateam2025gemma3technicalreport} can provide text-based summaries at lower latency, but text is limited in its ability to capture salient past details. Image-selection \citep{mark2026bpplongcontextrobotimitation} can provide more information, but is more computationally demanding at inference time.} %
{Therefore, this work proposes an alternative solution:}
\begin{AIbox}{}
\begin{quote}
    {Rather than using powerful models for deployment-time history curation, we can instead  leverage them for \textbf{training-time supervision} of a lightweight, latent representation of memory.}

\end{quote} 
\end{AIbox}

\iftoggle{arxiv}{\togglepar{ Our Contributions.}}
{}Concretely, we introduce  \textbf{workspace models}: a lightweight memory encoder for robotic policies that achieves high performance in tasks that require long-term memory, without the deployment-time computation  required by VLM-in-the-loop models.\footnote{Our approach is inspired by Global Workspace Theory in cognitive psychology, which posits that human beings maintain a bottleneck of information relevant to decision-making, memory, and perception \citep{bengio2019consciousnessprior, BAARS200545}}
Compared to these approaches, workspace models exploit a key asymmetry in robot learning: {test-time compute scaling must stay limited at deployment and training data is hard to scale, but training-time compute scaling can be exploited.}
Workspace models take advantage by leveraging a new  scaling axis we term \textbf{saliency-driven supervision}, where we expend inference compute on strong models to provide saliency labels at training time, which we use to train a compact encoder that captures all relevant historical information. 

We compare workspace models to a number of representative baselines, including a VLM-based frame selection, similar to \cite{mark2026bpplongcontextrobotimitation} across simulated and real-world tasks that require long-term memory for counting, spatial recall, and test-time adaptation. Across these tasks, workspace models enable effective memory with limited latency.  {Surprisingly}, \textbf{workspace outperforms test-time key-frame selection}, due to a number of key design advantages that we expose through careful ablations. Ultimately, we view the workspace model as both
\iftoggle{arxiv}
{
\begin{itemize}
    \item Opening the door to \textbf{novel architecture design} for long-horizon memory 
    \item A compelling proof of concept for \textbf{saliency-driven supervision} in domains where training data are scarce and/or computation is constrained at test-time
\end{itemize}
To this end, we stress that this manuscript's implementation of a workspace model---using keyframe patches for saliency-driven supervision---serves as only a preliminary instantiation of the workspace principle, and look forward to future work which investigates other modalities of supervision, including text and video, as well as applying the workspace principle across domains. 
}
{
opening the door to novel architecture design for long-horizon memory and serving as a compelling proof of concept for \textbf{saliency-driven supervision} in domains where training data are scarce and/or computation is constrained at test-time.
}

\begin{AIbox}{Workspace Models as {\color{\lightred}{Latent Harnesses}}}
In light of recent work demonstrating the power of frontier language models \citep{openai2026gpt6astra}, general-purpose robot control agents have emerged as an exciting new avenue where models observe the environment, maintain context, write programs, and invoke perception, planning, or control tools during execution. While these reasoning model capabilities are improving, the latency of their generations is also increasing. This poses a drawback for large models as reactive control policies which may require dynamic motion and precise control \citep{zhang2026unexpectedrobotpolicyearly}. Workspace Models suggest a complementary direction where we may not need to place expensive general-purpose reasoning into the deployment loop; instead, we can amortize components of agentic harnesses into latent modules. In our setting, a powerful model performs history curation during training, while a compact learned representation performs the corresponding memory computation at deployment.  \textbf{\emph{We view this as an initial example of a \colorbold{``latent harness''}}}: using frontier models to specify useful intermediate computation during training, then distilling this computation into representations suitable for fast downstream control.
\end{AIbox}

\section{Background: Training History-Based Robotic Policies}

\label{sec:related_work}
\newcommand{\history}{\mathtt{history}}
\vspace{.5em}

\iftoggle{arxiv}
{

}
{\vspace{-.1cm}}

\iftoggle{arxiv}
{
\newcommand{\Data}{D}
}
{
\newcommand{\Data}{\mathcal{D}}

}
\togglepar{Preliminaries. }We operate in the visual imitation learning setting. For a given task, we assume a dataset of $N$ expert demonstration trajectories, $\Data = \{ \tau_k\}_{k=1}^N$. Each trajectory is an ordered sequence of observations and actions $\tau_k = [(o_i, a_i)]_{i=1}^T$ with some episode length, $T$. Each observation contains the present image and proprioceptive state, $o_t=[I_t, x_t]$. The objective is to learn a receding horizon control policy $\pi(a_{t:t+H} \mid \history)$ where $H$ is the action prediction horizon and $\history$ is some conditioning variable depending on $o_{1:t}$. 

\begin{wrapfigure}{r}{0.39\textwidth}
\vspace{-3em}
\centering
\includegraphics[width=0.38\textwidth]{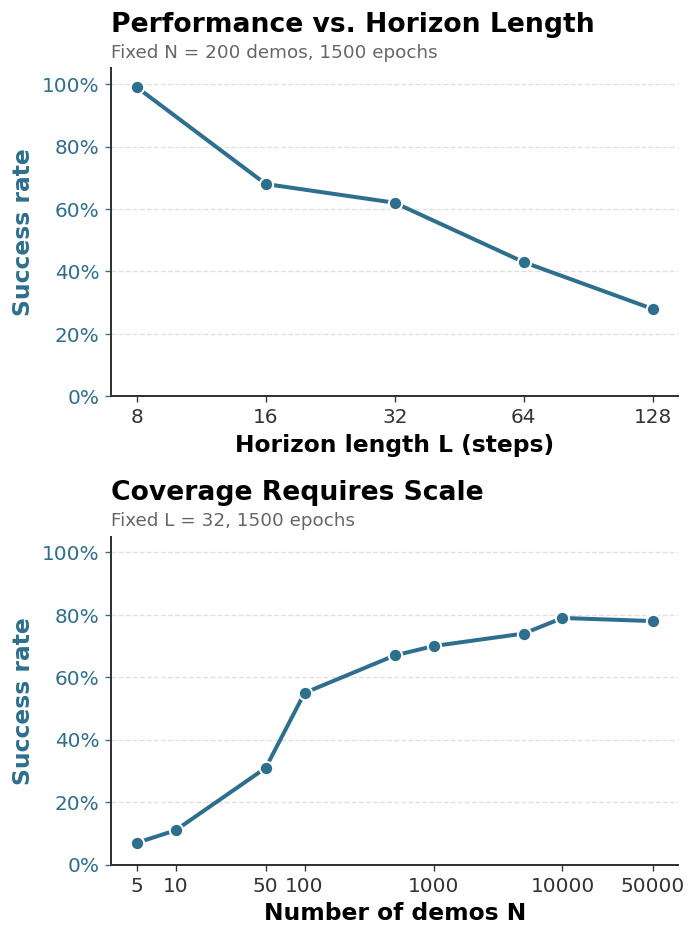}
\caption{Performance degrades with POMDP horizon.}
\label{fig:synthetic}
\end{wrapfigure}

\togglepar{History-conditioned Policy Learning. }Naively, one can select $\history = o_{1:T}$ to be all past observations. However, prior work has identified a number of challenges with this approach. Long histories introduce the risk of overfitting since spurious correlates inside irrelevant historical information can be used to learn ``shortcuts'' \citep{Geirhos_2020} for action prediction that do not generalize to the online rollout distribution. In imitation learning, this can surface as ``causal confusion'' \citep{de2019causal} or ``copycat behavior'' \citep{wen2020fightingcopycatagentsbehavioral} where the model exhibits poor performance by predicting actions from incorrect cues or simply copying previous actions, respectively.  
\iftoggle{arxiv}
{

}
{}
To illustrate this intuition, we train state-based policies with varying history requirements to navigate to a point goal, $g \in \{-1, +1\}$ at a timestep $L$, starting from $x_0=0$; however, $g$ is only shown to the agent once at some salient time $t_s \sim \mathcal{U}(1, L)$. Therefore, to succeed at reaching $g$ for some $L$, the agent must have an observation history of size $L$ or more. As seen in \Cref{fig:synthetic}, when training a policy conditioned on full history, the success rate degrades with increasing values of $L$, and even appears to saturate even as the number of demonstrations grows. As robot data is typically limited, the data-demands of naive history conditioning render the approach infeasible.

\togglepar{History Summarization for Robot Policy Learning.} Here, we describe past approaches to mitigating the challenges associated with history-conditioning; we defer an extended related work to \Cref{app:related}.
As mentioned in the Introduction,
recent methods use a large model online to curate a task-relevant subset of observations $o_{t_1},\dots,o_{t_k}$ or language summary $l_t$ to feed into a downstream learner. For example, \citet{torne2026memmultiscaleembodiedmemory} propose using a VLM to maintain a language memory and produce language subtasks for a downstream language-action model (VLA). \citet{sridhar2025memerscalingmemoryrobot} similarly build a high-level policy for VLA subtask generation but maintain a visual keyframe memory through a VLM and find that text-only memory does not capture all information necessary for planning. \citet{mark2026bpplongcontextrobotimitation} directly use an API-queried VLM to select keyframes to persistently feed into a Diffusion Policy \citep{chi2023diffusion} and handle latency through masking recent frames, however still report latency-related failure modes. Crucially, all of these methods rely on using large-scale VLMs during inference-time for memory selection. Querying these large models takes time, which introduces significant latency during the robot tasks. This compromises policy reactivity, harming success rates on more dynamic tasks \citep{mark2026bpplongcontextrobotimitation, sridhar2025memerscalingmemoryrobot}.  Moreover, large model queries incur significant financial cost, either because robots must ship with expensive GPUs, or pay for costly API queries whenever memory is needed. {Finally, the summary generation system must consistently provide all sufficient information to prevent online catastrophic failures; this introduces a trade-off between summary quality and generation latency/computational expense.} %

\section{Amortizing History Summarization via Workspace Models}
\label{sec:method}

\begin{figure}
    \centering
    \includegraphics[width=\linewidth]{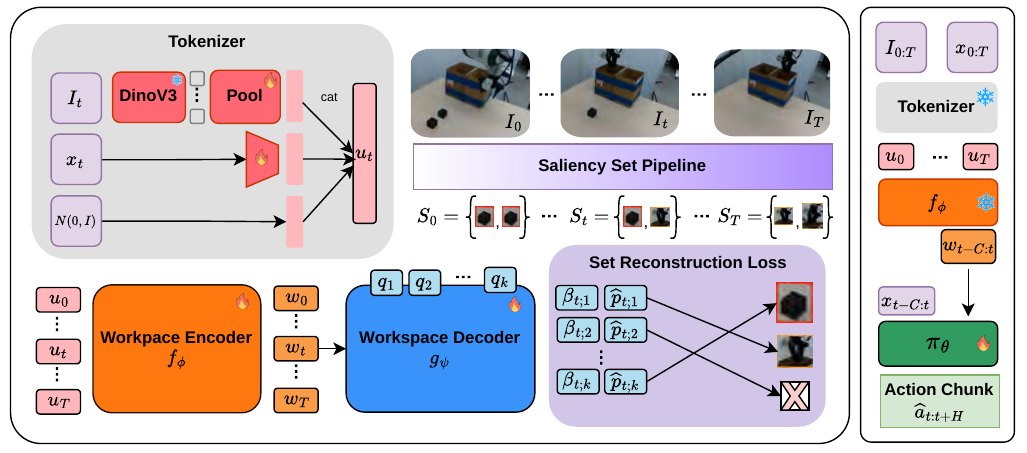}
    \caption{\textbf{Workspace Model Training (left):} Training involves tokenizing the image and proprioception through DinoV3 and a learned projection, respectively. These tokens are combined with an input slot at every timestep and fed to the workspace encoder to produce workspace tokens. Each workspace token, $w_t$, is supervised to reconstruct the set $S_t$ through a Hungarian matching set reconstruction loss on the workspace decoder outputs. \textbf{Imitative Model Training (right):} Images and proprioception are fed into the tokenization and workspace pipeline to produce a short history of workspace tokens. These are used as input to train a Diffusion Policy $\pi_{\theta}$, along with the raw proprioception, to output the future action chunks.}
    \label{fig:arch}
\end{figure}

\newcommand{\fwk}{f_{\phi}}
\newcommand{\gwk}{g_{\psi}}
\newcommand{\piwk}{\pi_{\theta}}
\vspace{.3em}

In this work, we introduce the \colorbold{workspace model} (\Cref{fig:arch}), a training pipeline and architecture that amortizes VLM queries during train-time to produce compact, latent summaries of salient history information. By design, our approach reduces the computational burden associated with querying powerful models at deployment time. 
\iftoggle{arxiv}
{

}
{
}
Workspace models train an autoregressive encoder $w_{t} = \fwk(o_{1:t})$ to produce a latent summary we call the \textbf{workspace token} $w_t$.
During training, we  use VLMs to curate a small \textbf{salient set} $S_t$ of image patches from past and current visual observations which a VLM deems salient for determining the optimal action $a_t$ (\Cref{sec:body_vlm_saliency}). Whereas prior work conditions on curated salient information at deployment, we instead train the encoder so that $w_t$ encodes all the information in the salient set $S_t$. This is accomplished by co-training $f_{\phi}$ with a decoder $g_{\psi}(w_t)$ and penalizing the set-distance between the latter's predictions and   the ground truth  $S_t$. Finally, we freeze the encoder $\fwk$, and learn a reactive, workspace-conditioned policy which produces actions from a short history of frozen workspace embeddings and proprioceptive states $\piwk(a_{t:t+H} \mid w_{t:t-C},x_{t:t-C})$. %
Via this approach, the reactive policy has access to a compact, latent history summary that contains all salient past information. %

\newcommand{\phattj}[1][j]{\hat{p}_{t;#1}}
\newcommand{\betatj}[1][j]{\beta_{t;#1}}
\newcommand{\qj}[1][j]{q_{#1}}
\newcommand{\vtj}[1][j]{v_{#1;t}}
\newcommand{\pti}[1][i]{p_{t;#1}}

\togglepar{Representing Salient Information via Key-Patch Reconstruction.}
Given that salient events in manipulation tasks are often localized in space and time, we choose to represent the salient set $S_t$ as 
an indexed set of DinoV3 \citep{simeoni2025dinov3} patch tokens from the images observed thus far. Concretely,   
$S_t = \{ \pti\}_{i=1}^{m}$ where $\pti$  consists of the Dino patch token at location $\ell_i$ chosen from the image at time $t_i$. We note that the Dino patches already contain an encoding of their patch location $\ell_i$. The saliency set is capped to be at most $m$ patches. In \Cref{sec:body_vlm_saliency} below, we describe how these patches are labeled and selected via prompted VLMs. We remark that this is only one possible representation of salient information, and leave exploration of other approaches to future work. 

\togglepar{Workspace Encoder Architecture.}
We choose to parameterize the workspace encoder,  $\fwk$ as a transformer with causal masking that receives a sequence of input tokens, $[(\bar{o}_{t'}, z_{t'})]_{t' \le t}$, where,  $\bar{o}_{t'} = \mathrm{tokenize}(o_{t'})$ are the Dino patch tokens of the observation in time $t'$. The $z_{t'}$ are inputs whose corresponding output-slots are the learned workspace tokens $w_{t'}$ (\cref{app:arch}), initialized to be $z_{t'} \overset{\mathrm{i.i.d.}}{\sim} \Normal(0,\eye)$. 
Applying the encoder to this sequence, $\fwk([(\bar{o}_{t'}, z_{t'})]_{t' \le t})$, produces the sequence of workspace tokens $w_{1:t}$.

\iftoggle{arxiv}{\vspace{.3em}}{}
\subsection{Workspace Model Training}
\iftoggle{arxiv}{\vspace{.3em}}{}
Importantly, we represent salient information as a \emph{set} $S_t$, which can be variable in size. Set structure necessitates specialized training and decoder design choices, which we describe here.  
We opt to train our decoder $\gwk$ following the DETR  framework \citep{carion2020endtoendobjectdetectiontransformers} for set reconstruction. Specifically, we parameterize our decoder $\gwk$ as a transformer to cross-attend learnable query tokens to a latent vector and produce $m$ ``slots'' (recall: $m$ is the maximum number of patches), each of which predicts information about a potential candidate set element together with a probability of whether that ``slot'' is occupied. 
Specifically,  $\gwk$ outputs $m$ pairs $\gwk(w_t)=\{( \phattj, \betatj)\}_{j=1}^{m}$ consisting of  (1)  patch reconstructions $\phattj$  and (2)    probabilities $\betatj$ that the slot $j$ is active at time $t$. The latter accounts for the salient set potentially containing fewer than $m$ elements.

\newcommand{\alphaocc}{\alpha_{\mathrm{active}}
}
\newcommand{\alphafeat}{\alpha_{\mathrm{feat}}
}
\newcommand{\crossent}{\mathrm{CrossEnt}}

\togglepar{Encoder/Decoder Training.}
 Training consists of two steps: a \emph{matching step} that computes a mapping $\sigma_t$ between available slots $j$ and patches in $S_t$, and a \emph{gradient step} on a reconstruction loss under that found matching. Given a binary $y \in \{0,1\}$, let $\crossent(\beta;y) := \I\{y = 1\}\log(1/\beta) + \I\{y = 0\}\log(1/(1-\beta))$ denote the standard cross-entropy loss.  Moreover, we define a ``null patch location'' $i = 0$, to which we map unoccupied slots $j$, and define $y(i) := \I\{i \ne 0\}$, indicating that $i$ is an active (non-null) position.

During the matching phase, we follow DETR and invoke the Hungarian algorithm \citep{hungarianalgo} to obtain a map $\sigma:[m]\to \{0,1,\dots,|S_t|\}$ between the $m$ available slots and entries of $S_t$ or null location $0$.  $\sigma_t$  is injective into $|S_t|$ (i.e. never maps two slots $j$ to the same $i \ge 1$, but can map many $j$ to $i = 0$),  and is computed by minimizing a cumulative matching cost $\sum_{\text{slots } j} c_t(j,\sigma_t(j))$, where the cost  $c_{t}(i,j) :=    y(i)\|\pti- \phattj\|^2 + \lambda_{0} \crossent(\betatj;y(i))$  penalizes mean squared error on non-null reconstructed patches and cross-entropy term on the indicators $y(i)$ that $i$ is ``active''. 
\newcommand{\cLactive}{\cL^{\mathrm{active}}}
\newcommand{\cLfeat}{\cL^{\mathrm{feat}}}

Given this matching\footnote{Importantly, we do not backpropagate gradients through the computation of $\sigma_t$} $\sigma_t$, we compute a gradient with respect to a weighted combination of  feature reconstruction (gated by slot $j$ being active) and  cross-entropy with the active labels $y(i)$:
\begin{align}
\cLfeat_{t} := \frac{1}{|S_t|} \sum_{j}y(\sigma_t(j))\|\phattj -\pti[\sigma_t(j)]\|^2, \quad \cLactive_t = \frac{1}{m}\sum_{j=1}^m \crossent(\betatj;y(\sigma_t(j)))
\end{align}

The full training objective is the sum of these per-timestep losses across the trajectory,
\begin{align}
\mathcal{L}
=
\mathbb{E}_{\tau}\left[\sum_{t=1}^{T}
\left( \lambda_1 \cLfeat_t + \lambda_2 \cLactive_t
\right)\right]
\end{align}
We train this model through mini-batch SGD using the AdamW optimizer and a linear warmup cosine decay learning rate scheduler. The training parameters are listed in \Cref{tab:hyperparameters}.

\togglepar{Workspace-Conditioned Policy.} Having learned the workspace encoder $\fwk$, we train the workspace action-head $\piwk$ to predict action-chunks \citep{zhao2023learningfinegrainedbimanualmanipulation} conditioned on a short history of workspace tokens and proprioceptive states:
$\piwk(a_{t:t+H} \mid w_{t-C:t}, x_{t-C:t})$. We implement $\piwk$ as a Diffusion Policy \citep{chi2023diffusion} further detailed in \Cref{app:dp}.

\begin{figure}[t]
    \centering
    \includegraphics[width=0.95\linewidth]{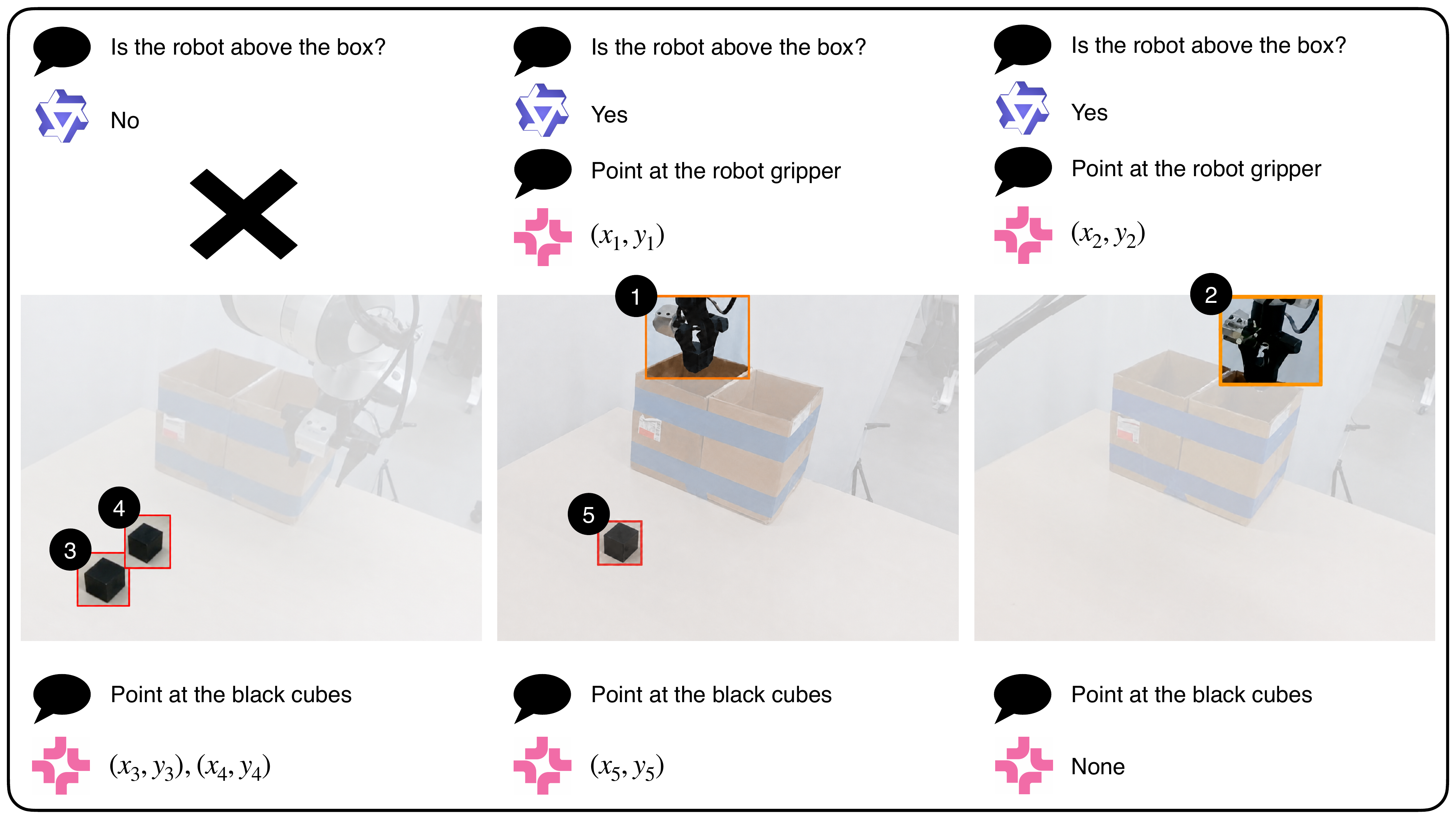}
    \caption{Our saliency-supervision pipeline which prompts Qwen (in purple) and MolmoPoint (in pink) to detect and locate events and build a growing memory. Orange boxes represent visual information which persists in the salient set over time. Red boxes represent ``transient'' information for that timestep only.}
    \label{fig:sal_pipeline}
\end{figure}

\subsection{VLM Saliency Labeling}
\label{sec:body_vlm_saliency}
We construct the salient set through a two-stage prompting depicted in \Cref{fig:sal_pipeline}. First, we use the Qwen3-VL-8B-Instruct \citep{bai2025qwen3vltechnicalreport} model to identify key time steps $t' \le t$ where salient events occur. Specifically, we provide Qwen with a set of classification questions to evaluate whether any given frame contains an event relevant to remember (e.g. is the salt put away?). To collect key-frames, we prompt Qwen with these questions on every frame of every trajectory. As a post-processing step, we filter out consecutive positive frames using a sliding window to remove redundant key-frames. The outcome is an ordered list of events to remember (e.g. salt was added to soup at time $t=50$, knife was put in drawer at $t=150$).

Subsequently, we use MolmoPoint \citep{clark2026molmopointbetterpointingvlms}, a VLM fine-tuned for open-vocabulary pointing, to extract salient patches at a regular frequency (refer to \Cref{app:prompt} for details) and for keyframes. At the regular frequency, MolmoPoint is given objects to persistently track (specified via prompt). For keyframes specifically, MolmoPoint is given a separate set of objects to track used only for keyframes. MolmoPoint then produces 2D coordinates for prompted objects for all keyframes and at the regular frequency.
We collect these keyframe-points from $t' \leq t$ and any persistent object points at $t$, effectively making the salient set contain both what to remember and what to focus on now. We then convert these points (each associated with a time) to DinoV3 patches by simply choosing the unique patch in time and space which overlaps with the point. The union of these constitutes the salient set $S_t$. We attach example prompts in \Cref{app:prompt}.

\section{Workspace Models Achieve High Task Success with Low Inference Latency}
\label{sec:result}
\begin{figure}
    \centering
    \includegraphics[width=\linewidth]{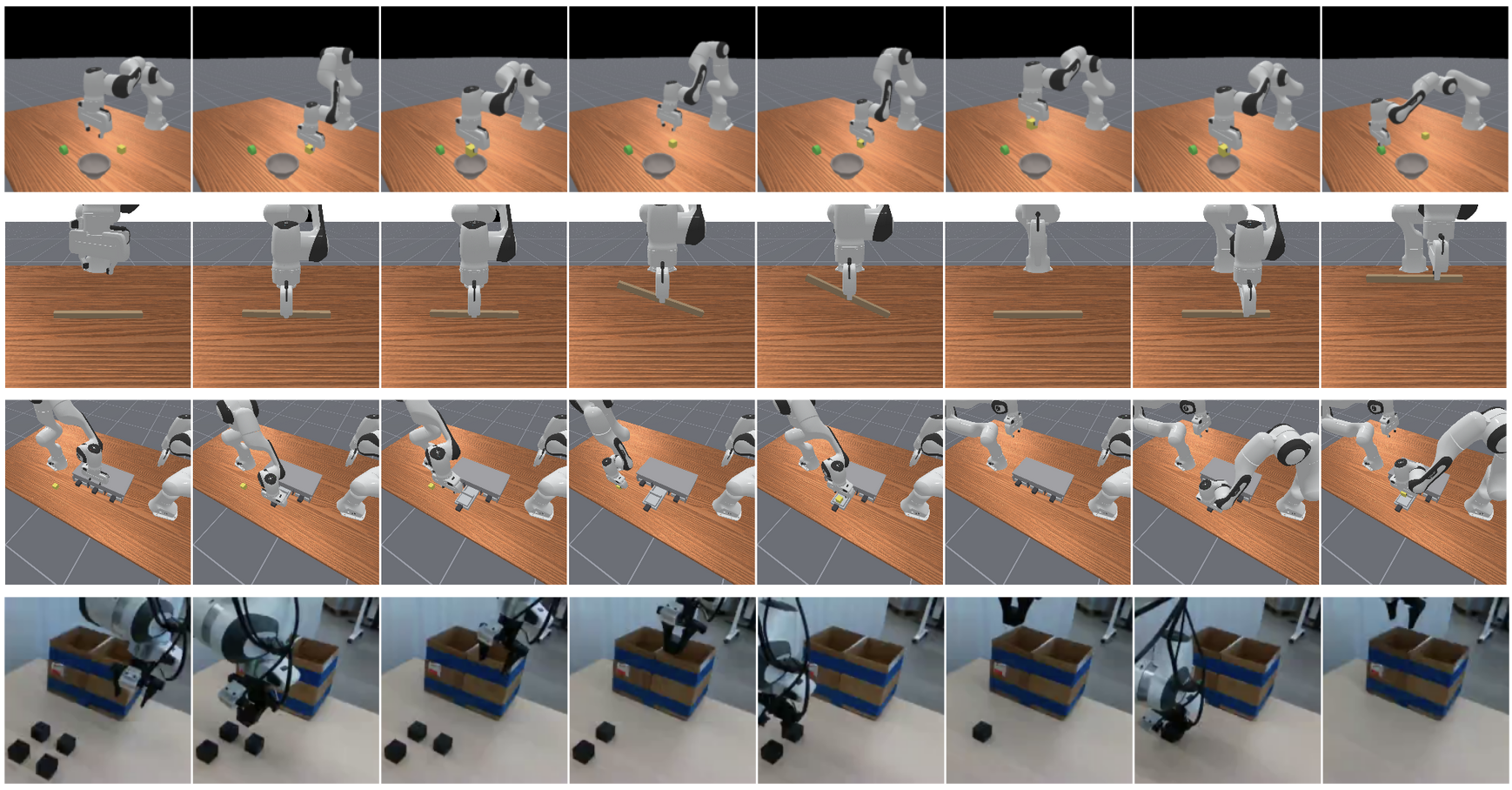}
    \caption{Every row depicts a representative demonstration for the tasks in order top-to-bottom: \cubedrop{}, \balance{}, \drawer{}, and \half{}. The first three rows are selected 3rd-person camera renderings for the trajectory. The fourth row is the global camera used for training.}
    \label{fig:filmstrips}
\end{figure}

In this section, we compare workspace models to a number of history-conditioning baselines. We show, across simulated and real-world environments, that workspace models enjoy low control latency while strongly \textbf{outperforming natural baselines in task success}. In the remainder of the section, we describe the baselines and tasks, and expose the core design decisions that contribute to the strong performance of workspace models relative to alternatives.

\colorpar{Baselines.} We compare workspace models, denoted $\wksp$, against three natural baselines: a vanilla Diffusion Policy, $\dpnc$, which uses the current and previous frame as input; a history-conditioned Diffusion Policy, $\dphist$ which conditions a sufficient history of frames by naively striding across previous observations (we find that full, unstrided history yields $0\%$ success); and finally, $\keyframe$, inspired by \citet{mark2026bpplongcontextrobotimitation}, which uses a VLM and a tuned prompt to select frames to add into Diffusion Policy context. We report baseline training and architecture details in \Cref{app:dp}. Importantly, \wksp{} \emph{uses very similar hyperparameters across all simulation and real-world environments}. Baselines are tuned per-task to steelman their performance. 

\begin{wraptable}{r}{0.45\textwidth}
\centering
\small
\caption{Latency comparison across baselines.}
\label{tab:latency}
\begin{tabular}{lrr}
\toprule
Method & Latency (ms) & Relative \\
\midrule
\dpnc & 41.3 & 1.00$\times$ \\
\dphist & 41.3 & 1.00$\times$ \\
\keyframe & 302 & 7.31$\times$ \\
\keyframe{}+Batch & 199.5 & 4.83$\times$ \\
\wksp & 94.9 & 2.3$\times$ \\
\wksp{}+Batch & 52.9 & 1.28$\times$ \\
\wksp{}+Batch+KV & 49.2 & 1.19$\times$ \\
\bottomrule
\end{tabular}
\end{wraptable}

\colorpar{Simulation and Hardware Setups.} We evaluate our methods against four memory-intensive tasks in both simulation and on real hardware. In simulation, we use ManiSkill3, built on SAPIEN \citep{taomaniskill3, chang2015shapenet, Mo_2019_CVPR, Xiang_2020_SAPIEN} to simulate our experiments on a Franka Research 3 robot \citep{franka_research3_manual}. Further details are provided in \Cref{app:exp_setup}. For real-world experiments, we use a Franka FR3 with an AgileX parallel jaw gripper with operational space control formulation~\citep{khatib1987unified}. Visual information is collected with a fixed Intel RealSense D435 global camera. 
Further details on teleoperation and controller information are provided in \Cref{app:exp_setup}. 

\subsection{Evaluation Tasks: Measuring Different Axes of Memory.} 
We test, in three simulated and one real task, three distinct memory-usage capabilities: \textbf{1)} counting under partial observability; \textbf{2)}  spatial recall and \textbf{3)} in-context adaptation from past experience. We describe the tasks at a high level below, and defer details to \Cref{app:exp_setup}. All baselines are trained via supervised behavior cloning from demonstrations.

To study the long-range counting capabilities, we design the $\cubedrop{}$ task in simulation, inspired by \citep{mark2026bpplongcontextrobotimitation}, in which a robot must place exactly 5 cubes into a bowl, where cubes leave the robot's line-of-sight when correctly placed.  We also evaluate on a hardware counterpart, \half{}, where the robot must place $K \in \{2,4\}$ cubes evenly into two boxes that occlude the visual camera. Both \cubedrop{} and \half{} necessitate memory, as partial observability obscures the running count from the current observation, while requiring that the memory does not interfere with precise execution (cube insertions).

Next,  \drawer{}  tests spatial recall. Here, another robot first opens and puts a cube away into one of three drawers, then closes it, and a separate robot (our policy) must open the drawer containing the stored cube.  Finally, we test in-context adaptation, in \balance{}. Here a robot is given two attempts to lift a bar with an unknown center-of-mass (CoM) so that it remains level. The demonstrations start with lifting from the middle first, revealing the CoM, then grasping at the CoM, resulting in a level lift. The learned policy must learn to do the same: use history to determine the CoM, and reattempt at the correct grip position. We show demonstration trajectories in \Cref{fig:filmstrips}.

\begin{figure}[t]
    \centering
    \includegraphics[width=\linewidth]{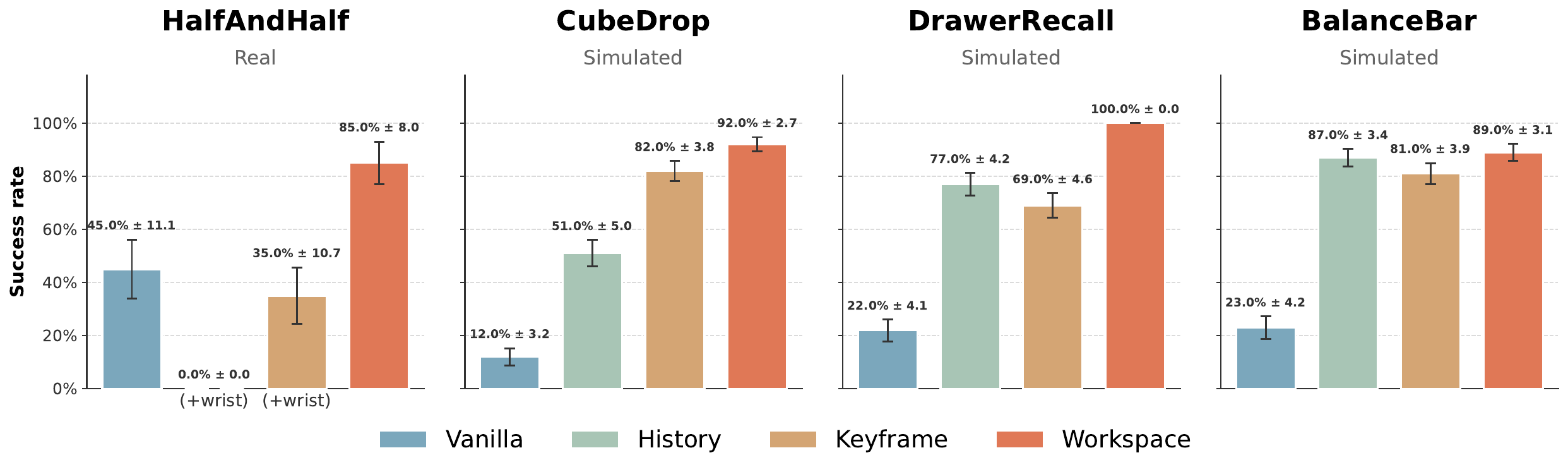}
    \caption{Success rates across 3 simulated (N=100) and 1 real task (N=20). For the real task, we use wrist cameras for the methods unable to pick the cubes correctly with a global camera only, denoted with ``(+wrist)''}
    \label{fig:real-results}
\end{figure}

\subsection{Workspace Models Successfully Amortize VLM Reasoning}
\iftoggle{arxiv}{\vspace{.3em}}{}
Recall our central motivation: amortizing slow and computationally-intensive VLM-in-the-loop history summarization into a lightweight encoder that can be queried efficiently at deployment. Before evaluating downstream policy performance, we verify that this amortization improves run-time.

Specifically, we measure the average amount of time to run the policy across the previous action chunk and replan the next chunk (shown in \Cref{tab:latency}). We measure relative to $\dpnc$ and find that $\wksp$ takes $\sim$1.2 times as long as a basic CNN perception stack as used in $\dpnc$ and $\dphist$. Meanwhile, $\keyframe$ is $\sim$5 times as costly, which would be significantly worse if using API call VLMs. We report numbers that leverage KV-caching and stacking observations between chunks into one batch. Together, these results confirm that workspace models achieve their core design goal - the expensive reasoning is done at train-time, leaving a lightweight and efficient encoder at deployment.

\subsection{Workspace models surprisingly outperform all 
baselines.} 
\vspace{-.5em}
\begin{wraptable}{r}{0.57\textwidth}
\centering
\vspace{-1.1em}
\small
\caption{Memory / Control failure rates across tasks.}
\label{tab:wrapped_failure_modes}
\begin{tabular}{lccc}
\toprule
Method & Bar (M/C) & Drawer (M/C) & Cube (M/C) \\ %
\midrule
\wksp{}     & 0\% / 100\%  & N/A         & 100\% / 0\% \\ %
\dpnc{}     & 88\% / 12\%  & 71\% / 19\% & 100\% / 0\% \\ %
\dphist{}   & 0\% / 100\%  & 50\% / 50\% & 100\% / 0\% \\ %
\keyframe{} & 0\% / 100\%  & 0\% / 100\% & 100\% / 0\% \\ %
\bottomrule
\end{tabular}
\end{wraptable}

\iftoggle{arxiv}{\vspace{.3em}}{}

Having established that workspace models successfully amortize VLM reasoning, we now evaluate whether this translates to downstream task performance. 
Figure \ref{fig:real-results} shows success rates across all four tasks. Workspace models achieve the highest success rate in every task, with a cross-task average of $91.5\% \pm 2.2$ significantly ahead of the next best baseline, $\keyframe$, at $66.8\% \pm 3.2$. Notably, this improvement holds in both simulation and the real world. This is our most \textbf{surprising finding}: $\wksp$ outperforms $\keyframe$ not only on latency (which was the original motivation), but also on success rate. Below, we dig into the mechanisms that lead to this performance gap.

\section{Why do Workspace Models Exhibit Better Task Performance?}

In this section, we account for the surprising finding that workspace models, despite being targeted at \emph{amortizing} key-frame lookup, actually \textbf{outperform} key-frame lookup. %

\subsection{Full Histories and Key-Frame History Both Induce Control-Failure}

We first begin by categorizing the failure modes exhibited by different methods as either memory-related errors, where the robot correctly executes a skill but targets the wrong mode (i.e. putting 3 blocks away and pressing done instead of 5), or control-related errors, where the robot fails to execute a manipulation skill (i.e. missing a grasp of a block). We share results of failures within simulation experiments in \Cref{tab:wrapped_failure_modes}, finding that $\dpnc{}$ unsurprisingly suffers from memory-related errors but rarely fails to execute an action mode. 

\colorpar{Weaknesses of Frame Stacking Methods.} The frame-stacking methods (i.e. $\dphist{}$ and \\
$\keyframe{}$) tend to contain good memory but fail more often with control. $\wksp{}$ seems to be able to balance both memory and control. Qualitatively, $\dphist$ and $\keyframe$ select the correct task mode, but fail on fine-grained motions. In \drawer{} for example, $\dphist$ often selects the correct drawer to open, but misses the handle slightly. $\keyframe$ exhibits these failure modes as well and sometimes freezes while opening the drawer.  In real-world experiments, we observe that failures of control precision can resemble mode-selection failures unrelated to the correct memory: for example, \keyframe{} and \dphist{} both attempt to pick in the middle of the two cubes or to grasp 8-10cm above the cube. This led us to form a hypothesis that the performance gap in Figure \ref{fig:real-results} stems from a common failure mode across both $\dphist$ and $\keyframe$ - adding more frames into context introduces generalization error that outweighs the benefit of richer history.

\begin{figure}
    \centering
    \includegraphics[width=0.49 \linewidth]{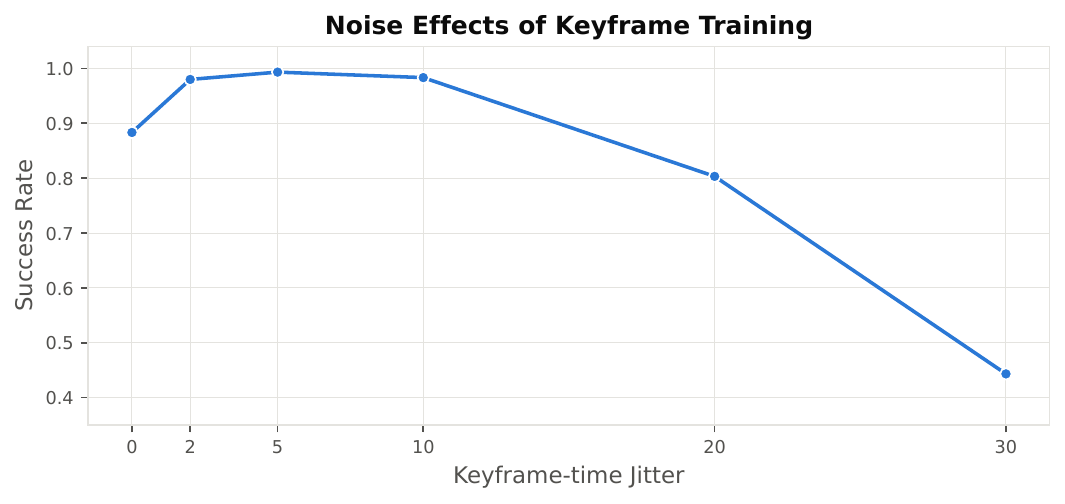}
    \hfill
    \includegraphics[width=0.49 \linewidth]{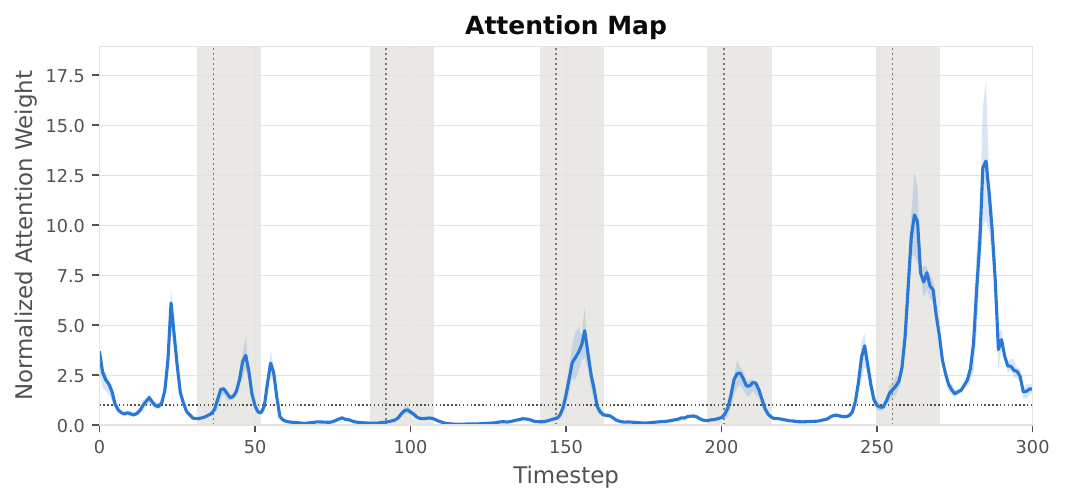}
    \caption{\textbf{Left:} Success as a function of noise from optimal timings. \textbf{Right:} Attention scores (average over 10 trajectories) measured relative to the average score at a given offset from the current timestep. Gray areas are segments of important events to remember}
    \label{fig:jitter_attn_big}
\end{figure}

\subsection{Keyframe-based methods introduce control aliasing.} 
Next, we identify a second failure mode:
 explicit frame selection methods (whether strided, as in \dphist{}, or VLM-based, as in \keyframe) create an \textbf{aliased} input structure during training that may be prone to distribution shift from their narrow training distributions. 
 
 To examine this, we train \keyframe{} with oracle keyframe timings at both training and deployment and find imperfect performance. Interestingly, adding moderate jitter to the training distribution of event times significantly improves performance as shown in \Cref{fig:jitter_attn_big}, but then falls as this training neighborhood is expanded too wide. This illustrates the balance between making the distribution of keyframes wide enough to prevent learning a brittle policy and not so wide that history coverage becomes slim and causes overfitting, as shown by \citet{mark2026bpplongcontextrobotimitation}.
The \keyframe{} method manages to introduce some jitter to counteract aliasing, as shown by its natural noise level ($\sigma \approx 10.1$) and strong performance with an oracle at test time ($\sim 0.96$). However, its success then relies on noisy VLM curation during train-time and perfect curation at deployment. Next, we show how $\wksp{}$ naturally remedies this by virtue of being a smooth representation.

\subsection{Why do Workspace Models Generalize Better?}
\iftoggle{arxiv}{\vspace{.3em}}{}

We now show that the workspace model exhibits smoothness in both its own training input, due to (1) training with full global attention over time, and (2) smoothness in its outputs (i.e. the policy inputs).

\begin{figure}[t]
    \centering
    \includegraphics[width=\linewidth]{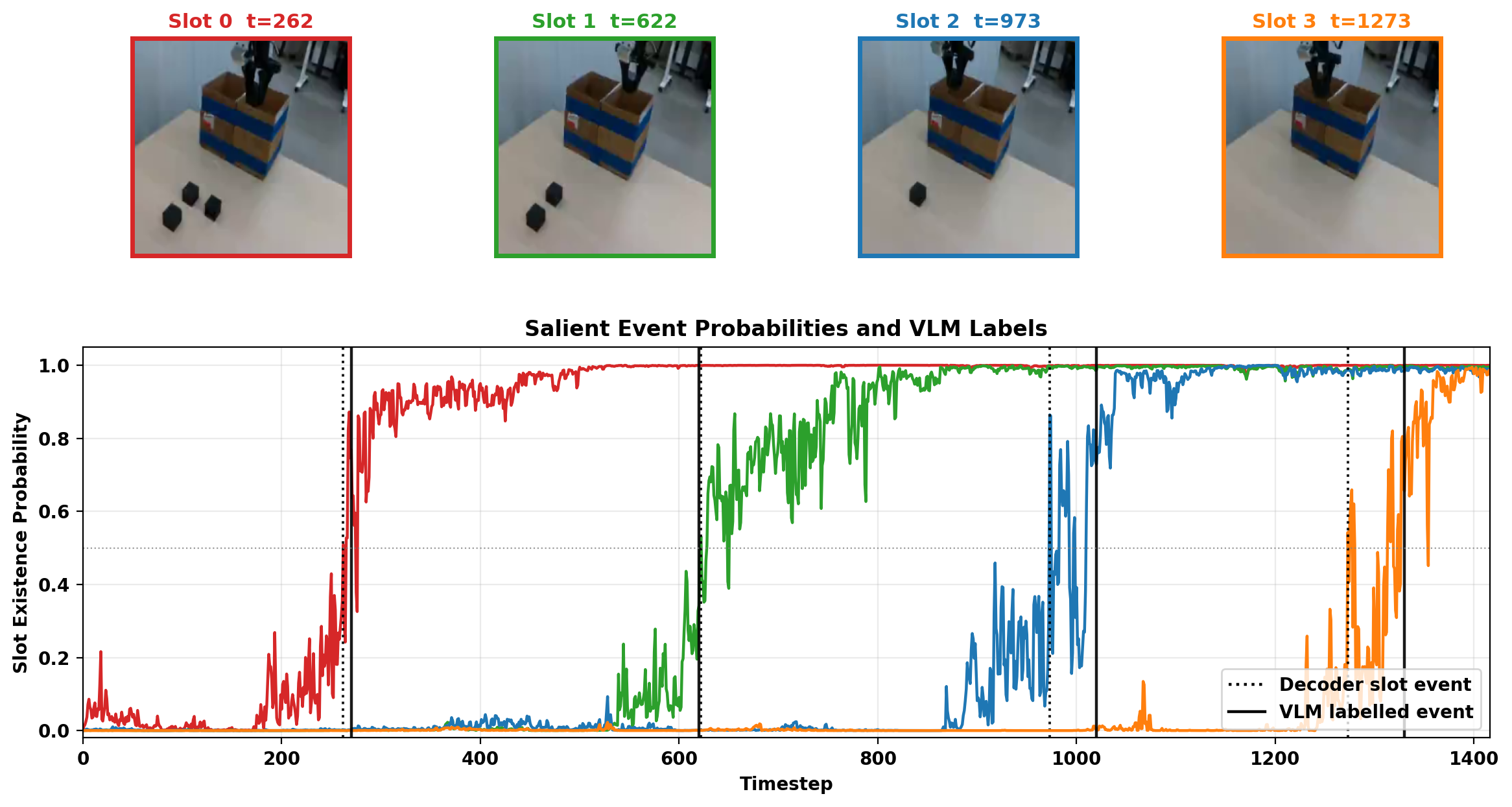}
    \caption{We show images and their timesteps which correspond to moments when a slot probability crosses a threshold of 0.5 (\textbf{top}). The moments where slots move above the threshold are ``decoder slot events'' shown with dotted lines. We also show the slot probabilities across time for every slot along with the VLM-supervised labels. As seen, decoder firings slightly diverge from the ground truth in order to match up to the event's general visual features, creating a smoothing effect (\textbf{bottom}). }
    \label{fig:decoder_firing}
\end{figure}

\colorpar{Benefits of global attention.} As evidence of (1), we identify that the workspace encoder itself attends across all available timesteps, learning to attend smoothly to the temporal sequence as opposed to paying full attention to discrete timesteps. In \Cref{fig:jitter_attn_big}, we show how the workspace encoder attends across adjacent salient times, smoothing out attention and mitigating sensitivity to aliasing effects as seen in its attention weights along with the decoder firing probabilities in \Cref{fig:decoder_firing}. 
 
\begin{figure}[!t] %
\centering

\begin{subfigure}[c]{0.65\textwidth}
    \centering
    \includegraphics[width=\linewidth]{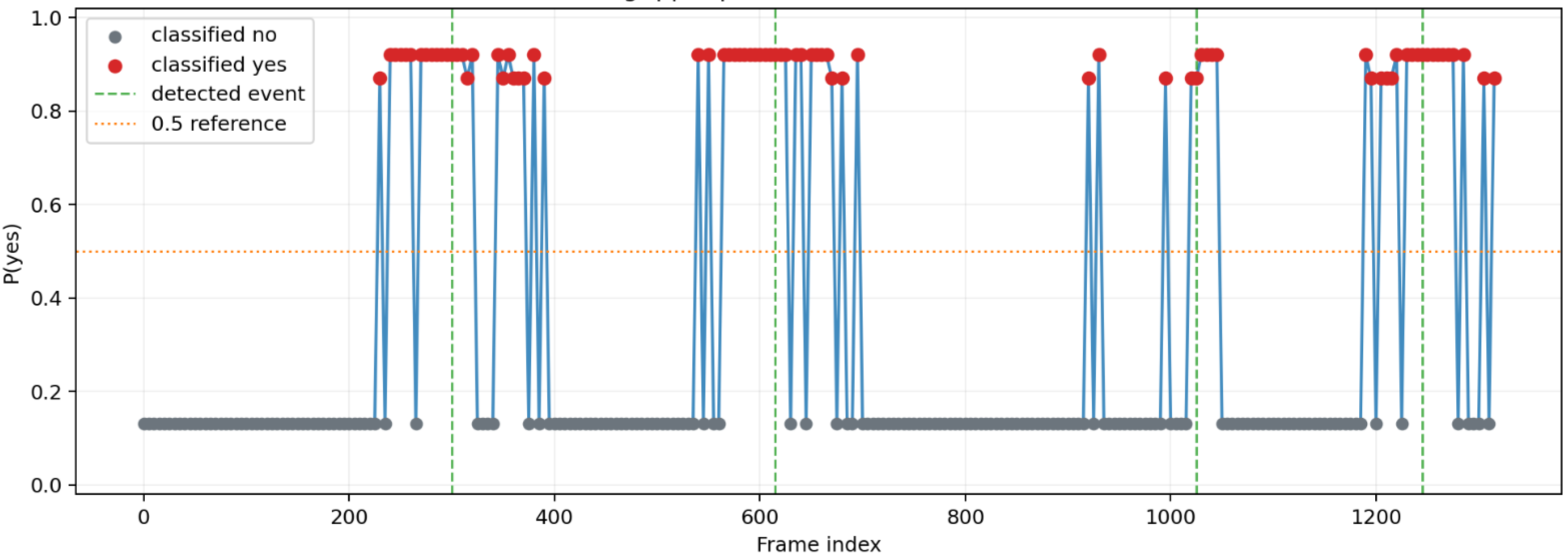}
\end{subfigure}
\hfill
\begin{subfigure}[c]{0.28\textwidth}
    \centering
    \includegraphics[width=\linewidth]{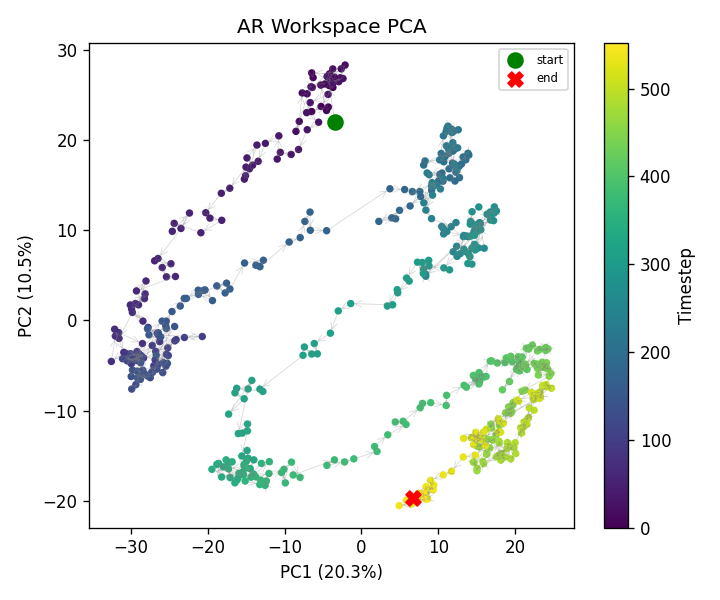}
\end{subfigure}

\caption{\textbf{Left:} A plot of the VLM detections of an event across time with the green lines representing the labels generated. \textbf{Right:} A PCA of the workspace token representations across a rollout.}

\label{fig:generalization}
\end{figure}

\colorpar{Smoothness in workspace outputs.} Towards (2), we show that workspace tokens (i.e. the workspace model outputs) are smooth from their PCA in \Cref{fig:generalization}. These  illustrate how tokens vary smoothly across time as well as their detected event triggers from the decoder. This smoothness is created by two major factors. First, our saliency pipeline creates smoothened \textit{median} labels as seen in \Cref{fig:generalization} since it has access to the full trajectory while labeling events as opposed to \keyframe{} which maintains a future-unaware labeling pipeline to match test-time labeling. Second, we naturally create smoothness through using an empirical risk minimization optimization across the dataset, creating an averaged encoding that must work robustly across many trajectories. This is seen in \Cref{fig:decoder_firing} where there exist semantic similarities between event frames and a smooth increase in probability over the event occurrence. With these results, we generally advocate for latent memory representations, and claim these characteristics of compressing history are especially significant to ensure downstream imitative policies are robust.

\section{Discussion}
\label{sec:discussion}

We present workspace models as a solution to the problem of long-horizon memory in robotic manipulation. By amortizing VLM reasoning at training time, we obtain a fast, robust encoder that generalizes better than VLM-in-the-loop alternatives.  The specific instantiation we propose (saliency sets as image patches, a DETR-style decoder) is one way to realize this idea, and many alternatives are worth exploring as well as other forms of VLM supervision. For instance, future work could extend the method to multi-camera streams, encode language reasoning traces, or incorporate dynamics prediction. 

More broadly, we hope this work encourages that perspective. The core principle extends beyond robotics and memory -  anywhere a powerful foundation model is queried repeatedly at test-time to solve a structured subtask, there may be an opportunity to amortize that reasoning into a faster and more robust learned module. And as we have shown, amortization can lead to even stronger results than stronger models in the loop.

\togglepar{Limitations.} While the workspace model works well, it relies on a fully autoregressive transformer encoder. This quadratically scales computation, and can become cumbersome at very long sequence lengths. To ameliorate this, we can train workspace models to be recurrent, block diagonal in their attention masks to reduce overhead, or employ KV-caching. Workspace models also receive supervision on what visual information is salient through a VLM which may not be fully grounded in determining which parts of an image are relevant to control. Reinforcement learning can provide an option for grounding in task reward. Lastly, we do not compare to hierarchical methods utilizing language-summaries and VLAs, but this would be interesting to examine along with how workspaces can be combined with generalist policies in the multi-task setting.

\section{Acknowledgments}
\label{sec:acknowledgments}
We want to express our gratitude to Younghyo Park, Ryan Bahlous-Boldi, and Antonia Bronars for relevant commentary about the work, along with the Improbable AI Lab community for fruitful discussions. This research was financially supported by the Ministry of Trade, Industry, and Energy (MOTIE), Korea, under the "Global Industrial Technology Cooperation Center program" supervised by the Korea Institute for Advancement of Technology (KIAT). (Grant No. P0028435). This material is based upon work supported
by the National Science Foundation Graduate Research Fellowship under Grant No. (NSF grant 2141064).

\section{Author Contributions}
\label{sec:author_contributions}
\textbf{Nitish Dashora} co-developed the project direction, architecture, experimental design, hardware experiments, paper writing, and code development

\noindent \textbf{Douglas Chen} contributed to writing, experimental design, and code development

\noindent \textbf{Idan Shenfeld} co-developed the project direction, contributed to writing and experimental design, and provided conceptual guidance

\noindent  \textbf{John Marangola} contributed to hardware experiments

\noindent  \textbf{Pulkit Agrawal} played a role in paper writing and high-level advising

\noindent \textbf{Max Simchowitz} co-developed the project direction, architecture, experimental design, and played a significant role in paper writing

\bibliographystyle{plainnat}
\bibliography{refs}
\newpage
\tableofcontents 

\newpage
\appendix
\crefalias{section}{appendix} 
\crefalias{subsection}{appendix} 

\section{Extended Related Work}
\label{app:related}

\textbf{Generalization in Robotics.} We desire a system which can be conditioned on historical information to enable capabilities like in-context learning, adaptation, or memory-intensive reasoning. But in many settings, exposing a learning system to more information can actually hurt performance. \citet{Geirhos_2020} describe shortcut learning, the phenomenon where deep learning systems can learn ``shortcut'' strategies to achieve the training objective that do not generalize to other conditions. This is connected to causal confusion seen in imitation learning \citep{de2019causal} or observational overfitting in RL \citep{song2019observationaloverfittingreinforcementlearning} where access to more information seemingly hurts performance by giving more potential irrelevant cues to learn from, thus hurting generalization. When policies are exposed to observational histories, this confusion can also show up as ``copycat'' behavior \citep{wen2020fightingcopycatagentsbehavioral}, where the learner exploits short-range correlations in demonstration trajectories rather than inferring the latent state that should drive action selection. This makes learning policies that require history especially difficult without auxiliary temporal objectives \citep{torne2025learninglongcontextdiffusionpolicies} or some form of information selection \citep{mark2026bpplongcontextrobotimitation, sridhar2025memerscalingmemoryrobot}. There are a handful of classic approaches to manage information selection.  

\textbf{Information Selection.} The idea of information filtering has roots in classic machine learning, statistics, and information theory. Early contributions involved implicit feature selection through the Lasso penalty \citep{10.1111/j.2517-6161.1996.tb02080.x} or explicit statistical testing \citep{10.5555/832245.832359}. Other methods use information-theoretic objectives \citep{dt-article, 10.5555/2503308.2188387} or correlative measures \citep{yu2003feature}. However, these methods often break down in deep learning with high-dimensional data. A popular line of work for studying compression of information and task-relevance is the information bottleneck (IB) design choice \citep{tishby2015deeplearninginformationbottleneck}, where the architecture design forces compression of information in the middle of the neural network. However, models still fall prey to not knowing what may be falsely task-relevant without having enough data \citep{de2019causal, Geirhos_2020}, which is exacerbated in high-dimensional settings. Moreover, IB lacks a direct analog for modern diffusion-based pipelines. Some newer methods involve learning attention maps, but they do not learn explanations for predictions, and are uncorrelated with feature importance \citep{jain2019attentionexplanation}, incurring quadratic costs. Other methods involve more active selective processing \citep{mnih2014recurrentmodelsvisualattention} but require expensive RL tuning. So, many practitioners have turned to using human-like general priors for determining what information is useful to maintain for a task. These usually leverage some foundation model, particularly a vision-language model (VLM). 

\textbf{VLM Guidance.}
One way to leverage a VLM for this is to supervise where the policy should attend. Some work has shown that human gaze can serve as an auxiliary signal for action planning \citep{land1999gaze} and robust imitation learning \citep{saran2021efficientlyguidingimitationlearning, banayeeanzade2025gabrilgazebasedregularizationmitigating}, but collecting gaze or human-saliency is expensive \citep{liang2024visarlvisualreinforcementlearning}. Therefore, VLMs have been used to provide guidance through different means such as saliency maps \citep{gong2025autofocusilvlmbasedsaliencymaps}, direct supervision \citep{blank2024scalingrobotpolicylearning}, or planning \citep{stone2023openworldobjectmanipulationusing, duan2024manipulateanythingautomatingrealworldrobots}. Furthermore, a growing line of work concerns grounding VLMs in order to provide spatial information that can be leveraged by downstream controllers to replace narrowly-trained perception systems \citep{yuan2024robopointvisionlanguagemodelspatial, clark2026molmopointbetterpointingvlms}. However, these VLM systems don't directly address the history compression problem we see when long contexts are required. For this problem, \citet{mark2026bpplongcontextrobotimitation} study using VLMs during execution to select keyframes that correspond to behaviorally salient events.  \citet{sridhar2025memerscalingmemoryrobot} and \citet{torne2026memmultiscaleembodiedmemory} use multimodal memory to produce ongoing language summaries to prompt a VLA with. These techniques all require heavy in-the-loop VLM planning, pointing towards latent memory as a potential solution. Our method implements a latent form of memory resembling Cartridges \citep{eyuboglu2025cartridgeslightweightgeneralpurposelong}, KV-caches distilled from corpora offline to amortize attention for reasoning. Workspace Models similarly distill VLM history curation into a latent embedding for memory-intensive planning. 

\newcommand{\MSE}{\mathrm{MSE}}
\section{Architecture and Training Details}
\label{app:arch}
\subsection{Diffusion Policy Details}
\label{app:dp} 
\begin{table}[h]
\centering
\caption{Hyperparameters for Diffusion Policy Training.}
\label{tab:diffusion_hyperparameters}
\begin{tabular}{lll}
\toprule
\textbf{Component} & \textbf{Hyperparameter} & \textbf{Value} \\
\midrule
\multirow{3}{*}{Training}
  & \texttt{--total-iters}                  & 500{,}000 \\
  & \texttt{--batch-size}                   & 128 \\
  & \texttt{--lr}                           & $1\times10^{-4}$ \\
\midrule
\multirow{7}{*}{Architecture}
  & \texttt{--obs-horizon}                  & 2 \\
  & \texttt{--act-horizon}                  & 8 \\
  & \texttt{--pred-horizon}                 & 16 \\
  & \texttt{--diffusion-step-embed-dim}     & 64 \\
  & \texttt{--unet-dims}                    & [64, 128, 256] \\
  & \texttt{--n-groups}                     & 8 \\
  & \texttt{--num-denoising-steps}          & 100 \\
\midrule
\multirow{1}{*}{Data}
  & \texttt{--control-mode}                 & \texttt{pd\_ee\_pose} \\
\bottomrule
\end{tabular}
\end{table}

We train imitative policies by approximating the conditional action distribution $p(A_t | O_t)$ through a diffusion model, as done in Diffusion Policy \citep{chi2023diffusion}. This is done by sampling a true data sample, $A^0_t$ from $\mathcal{D}$, iteratively corrupting it with noise through a schedule parameterized by $\alpha$, $\gamma$, and $\sigma^2$, and learning a network, $\epsilon_{\theta}$, to predict the corruption noise through the following loss:
$$
\mathcal{L}_{\mathrm{action}} = \MSE(\epsilon_k, \epsilon_{\theta}(A_t^0 + e_k, O_t, k))
$$
To sample from this learned distribution, we first sample a noisy $A^k_t \sim \mathcal{N}(0, I)$ and denoise $k$ times to produce an uncorrupted sample $A^0_t$. Denoising is done through the following equation:
$$
A_t^{k-1} = \alpha(A^k_t - \gamma \epsilon_{\theta}(A^k_t, O_t, k) + \Normal(0, \sigma^2I))
$$
Our noise-prediction network, $\epsilon_{\theta}$ is implemented as a 1D U-Net with FiLM conditioning \citep{perez2017filmvisualreasoninggeneral, ronneberger2015unetconvolutionalnetworksbiomedical} where $O_t$ linearly modulates the activations through the U-Net. We append training details for the \wksp{} and \dpnc{} models in \Cref{tab:diffusion_hyperparameters}.

\subsection{Workspace Model Details}
For training, we use one seed and utilize an 80/20 split. We choose the lowest-validation-loss model and embed the full dataset with that model for imitative training. We now specify model details in this section, particularly how information is fed into the workspace encoder. At every timestep, an image $I_t$ and proprioceptive state $x_t$ are received. We convert $I_t$ into patches through DinoV3 and use a single layer cross-attention pooling mechanism. This converts all patches into 1 token (through a learned fixed query token). We use a simple MLP to map $x_t$ into a proprioceptive token and concatenate it with the pooled image token, giving a tokenized observation function $\bar{o}_{t} = \mathrm{tokenize}(o_{t})$ where $o_{t} = [I_t, x_t]$. We use a sequence of these tokenized inputs, each with its own padding token, $z$, to generate a sequence of workspace tokens in one forward pass. The outputs corresponding to the padding inputs are the workspace tokens. During training we use causal masking so that during runtime we do not require the future when producing the current workspace token. In practice, we use a learned positional embedding within every timestep (across the image, proprioceptive, and padding tokens), and a fixed sinusoidal positional embedding across all timesteps.
\begin{table}[t]
\centering
\caption{Hyperparameters for Workspace Training.}
\label{tab:hyperparameters}
\begin{tabular}{llllll}
\toprule
\textbf{Component} & \textbf{Hyperparameter} & \textbf{CD} & \textbf{DR} & \textbf{BB} & \textbf{HnH}\\
\midrule
\multirow{5}{*}{Backbone}
  & Model                       & \texttt{DCT} & \texttt{DCT} & \texttt{DCT} & \texttt{DVB} \\
  & Number of patches           & 49 & 49 & 49 & 196 \\
  & Patch dimension             & 768 & 768 & 768 & 768 \\
  & Register tokens             & 0 & 0 & 0 & 4 \\
  & Frozen                      & True & True & True & True\\
\midrule
\multirow{4}{*}{Pooler}
  & Pool tokens                 & 1 & 1 & 1 & 1 \\
  & Layers                      & 1 & 1 & 1 & 1 \\
  & Attention heads             & 4 & 4 & 4 & 4 \\
  & Dropout                     & 0.1 & 0.1 & 0.1 & 0.1 \\
\midrule
\multirow{7}{*}{Encoder}
  & Hidden dimension            & 768 & 512 & 512 & 768 \\
  & Layers                      & 2 & 3 & 3 & 2 \\
  & Attention heads             & 4 & 4 & 4 & 4 \\
  & MLP ratio                   & 2.0 & 2.0 & 2.0 & 2.0 \\
  & Dropout                     & 0.1 & 0.1 & 0.1 & 0.1 \\
  & Slots                       & 8 & 8 & 8 & 8 \\
  & Workspace tokens            & 1 & 1 & 1 & 1 \\
\midrule
\multirow{7}{*}{Decoder}
  & Layers                      & 2 & 2 & 2 & 2 \\
  & Attention heads             & 4 & 4 & 4 & 4 \\
  & MLP ratio                   & 2.0 & 2.0 & 2.0 & 2.0 \\
  & Dropout                     & 0.1 & 0.1 & 0.1 & 0.1 \\
  & Trunk hidden dim            & 768 & 768 & 768 & 768 \\
  & Trunk layers                & 2 & 2 & 2 & 2 \\
  & Feature dim                 & 768 & 768 & 768 & 768 \\
\midrule
\multirow{4}{*}{Loss}
  & Existence loss weight       & 1.0 & 0.01 & 0.01 & 0.01 \\
  & Feature loss weight         & 1.0 & 2.0 & 2.0 & 1.0 \\
  & Existence cost              & 1.0 & 1.0 & 1.0 & 1.0 \\
  & Feature cost                & 0.5 & 1.0 & 1.0 & 1.0 \\
\midrule
\multirow{6}{*}{Training}
  & Batch size                  & 32 & 24 & 32 & 8 \\
  & Learning rate               & $1\times10^{-4}$ & $1\times10^{-4}$ & $1\times10^{-4}$ & $1\times10^{-4}$ \\
  & Weight decay                & 0.01 & 0.01 & 0.01 & $1\times10^{-4}$ \\
  & Warmup steps                & 250 & 250 & 250 & 1000\\
  & Max steps                   & 5000 & 12500 & 15000 & 25000 \\
  & Gradient clip               & 1.0 & 1.0 & 1.0 & 1.0\\
\bottomrule
\end{tabular}
\end{table}

We include hyperparameters for training our model in \Cref{tab:hyperparameters} with abbreviations CD (\cubedrop), DR (\drawer), BB (\balance), and HnH (\half). We also shorten \texttt{dinov3-convnext-tiny-pretrain-lvd1689m} as DCT and \texttt{dinov3-vitb16-pretrain-lvd1689m} as DVB.

\section{Prompting}
\label{app:prompt}
\subsection{Event Detection}
Here, we discuss the details of the VLM labeling pipeline. Recall that the workspace token contains information about both the past and present, \textbf{compressing both temporally and spatially}. Therefore, at a given time step, we distinguish between two different types of patches in our salient sets. At every time step, we can have \textbf{event} and \textbf{transient} patches. Event patches are specialized to an event, are chosen at the time of the event, and remain in the salient set. Transient patches are relevant in the present only and exist in the salient set for the current time. 

This leads us to the first set of parameters for the labeling pipelines. We have 4 prompts: \texttt{task\_desc-} \\\texttt{ription}, \texttt{event\_detection}, \texttt{event\_patch}, and \texttt{transient\_patch}, all working together to label events and patches. We primarily use two models, \texttt{Qwen3-VL} for event labeling, and \texttt{MolmoPoint-8B} for pointing (which is then used for patch labeling). We pass in prompt templates (\cref{fig:prompt_templates}) alongside the current RGB frame to the labeling models. The event labeling returns either ``yes'' or ``no'' for whether an event has been detected. The point labeling returns a list of \texttt{(pixel\_x, pixel\_y)} points in the RGB frame's pixel space, which we normalize to $[0, 1]$ and ultimately convert into patch indices. We also have 2 other shared parameters: \texttt{sample\_rate}, \texttt{max\_events}. The sampling rate is the frequency at which we run event labeling. Meanwhile, the max event count controls the maximum number of events; we detail the effect this has on post-processing for each pipeline in \Cref{app:post-processing}

\subsection{Post Processing}
\label{app:post-processing}
For the $\wksp$ labeling pipeline, all positive frames are first sorted by their time step. We then group them into segments where a new segment starts only when the gap from the previous positive frame is larger than \texttt{min\_event\_separation\_steps}. Then the middle position frame in each segment is chosen as the event. We do a final post processing step: if we have more than \texttt{max\_events}, we take the first \texttt{max\_events} events. For $\wksp$ we also track transient patches every \texttt{tracking\_interval} frames.

For the $\keyframe$ labeling pipeline, we only start classifying after \texttt{ignore\_before} and we treat any timestep on a rising edge, where the current classification is ``yes'' and the previous classification is ``no'', as a keyframe. After detecting a keyframe, we skip ahead by \texttt{cooldown} and continue sampling every \texttt{sample\_rate} after as usual. We store the keyframes in a FIFO buffer; when we go over \texttt{max\_events}, we evict the oldest keyframe and add in the new keyframe.

\begin{figure}[t]
\centering
\begin{minipage}{0.95\linewidth}

\begin{promptbox}{Event Prompt}
\ttfamily\small
This is frame \{timestep\} from a robot demonstration of the task
\{task\_description\}. \{event\_prompt\}. Look ONLY at this single frame.
Answer with a single word: yes or no.
\end{promptbox}

\begin{promptbox}{Point Prompt}
\ttfamily\small
Point to the \{transient/event patch\}
\end{promptbox}

\end{minipage}
\caption{Prompt templates used for event detection and point prediction.}
\label{fig:prompt_templates}
\end{figure}

\begin{table}[t]
\centering
\caption{Parameters for VLM Prompting.}
\label{tab:prompt_parameters}
\begin{tabular}{llllll}
\toprule
\textbf{Component} & \textbf{Hyperparameter} & \textbf{CD} & \textbf{BB} & \textbf{DR} & \textbf{HnH} \\
\midrule
\multirow{1}{*}{Event}
  & Model & \texttt{Qwen3-VL} & \texttt{Qwen3-VL} & \texttt{Qwen3-VL} & \texttt{Qwen3-VL} \\
\midrule
\multirow{1}{*}{Transient}
  & Tracking Interval & 2 & 2 & 2 & N/A \\
\midrule
\multirow{1}{*}{\wksp}
  & Sample Rate & 5 & 4 & 5 & 5 \\
  & Max Events & 5 & 1 & 1 & 4 \\
  & Min Event Separation & 20 & 16 & 15 & 75 \\
\midrule
\multirow{1}{*}{\keyframe}
  & Sample Rate & 5 & 2 & 15 & 5 \\
  & Cooldown & 30 & 96 & 1300 & 300 \\
  & Ignore Before & 0 & 22 & 100 & 50\\
\bottomrule
\end{tabular}
\end{table}

\begin{promptbox}{CubeDrop Prompts}
\ttfamily\small
{\normalfont\small\bfseries Event Detection Prompt}\\
Is the black gripper above the gray bowl?

\medskip
{\normalfont\small\bfseries Event Point Prompt}\\
gray bowl

\medskip
{\normalfont\small\bfseries Transient Tracking Point Prompt}\\
yellow block
\end{promptbox}

\begin{promptbox}{BalanceBar Prompts}
\ttfamily\small
{\normalfont\small\bfseries Event Detection Prompt}\\
Is the gripper lifting the brown bar straight up from the middle while the bar is still close to the table surface?

\medskip
{\normalfont\small\bfseries Event Point Prompt}\\
brown balance bar

\medskip
{\normalfont\small\bfseries Transient Tracking Point Prompt}\\
brown balance bar
\end{promptbox}

\begin{promptbox}{DrawerRecall Prompts}
\ttfamily\small
{\normalfont\small\bfseries Event Detection Prompt}\\
Is the left robot gripper grasping the yellow block with an open drawer in the scene?

\medskip
{\normalfont\small\bfseries Event Point Prompt}\\
the drawer that the left robot gripper is opening

\medskip
{\normalfont\small\bfseries Transient Tracking Point Prompt}\\
yellow block
\end{promptbox}

\begin{promptbox}{HalfAndHalf Prompts}
\ttfamily\small
{\normalfont\small\bfseries Event Detection Prompt}\\
Is the black robot gripper above the brown box?

\medskip
{\normalfont\small\bfseries Event Point Prompt}\\
robot gripper

\medskip
{\normalfont\small\bfseries Transient Tracking Point Prompt}\\
black cubes
\end{promptbox}

\section{Experimental Setup and Details}
\label{app:exp_setup}
\subsection{Simulation}
ManiSkill3 is configured to run at 100 Hz. All data collection in simulation is scripted with simple linear motion planners with privileged state information. We run our policies and data collection at 20Hz with PD control on a 7D action including end-effector pose and gripper state. 

\textbf{\cubedrop} In this task, the robot must place exactly 5 cubes into a bowl; however, once something is placed in the bowl, it becomes invisible. We randomly re-spawn blocks after they are placed in the bowl to emphasize the need for present-time perception as well. Once 5 cubes are added, the robot must press a green button to confirm it is done. The robot succeeds if and only if it presses the green button after having exactly 5 cubes in the bowl. We use 272 trajectories for training. The main failure modes we observe with \dpnc{} involve not pressing this green button at the right time; however, it does exhibit the strongest manipulation capabilities, always quickly and effectively picking the cubes. This supports the hypothesis regarding how lower observation horizons allow for better function fitting. When adding other frames, \dphist{} and \keyframe{} sometimes miss grasps, with \keyframe{} alleviating the fitting issues as the input frames are more consistent.

\textbf{\drawer} In this task, the robot observes a secondary robot open a drawer, pick up a cube, place it into the drawer, and then close it. This is done such that it is impossible for the primary robot to know which drawer the cube is within after the drawer is closed. The robot's task is to then open the correct drawer. We utilize 272 demonstrations for training

\begin{wrapfigure}{r}{0.35\textwidth}
\centering
\includegraphics[width=0.34\textwidth]{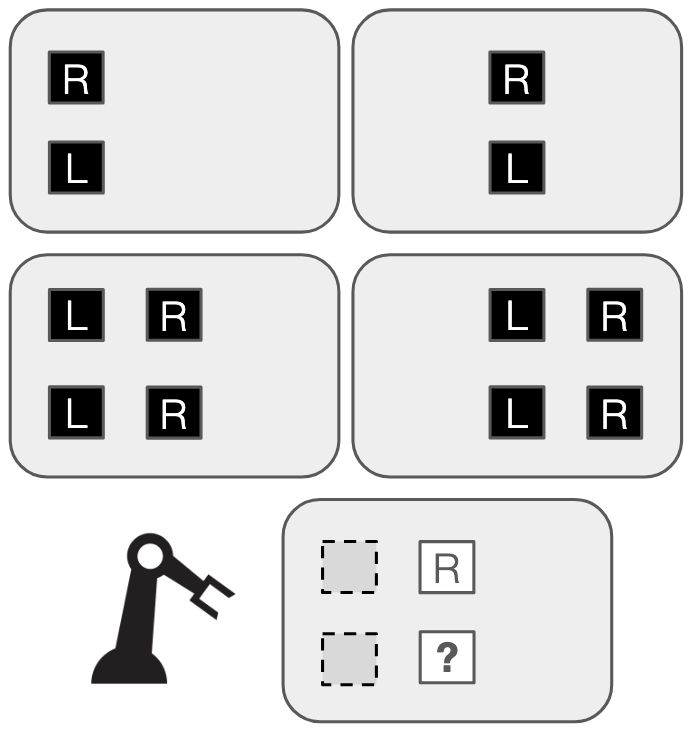}
\caption{Each black square represents a cube, and the letter represents where the robot must put it (either the right or left box). The test configuration requires memory of the past}
\label{fig:teleop}
\end{wrapfigure}

\textbf{\balance} This task involves lifting a long bar such that it is level (e.g., zero degrees of tilt about the midpoint). However, the bar has one of 3 centers of mass (CoM). The demonstrations begin with an attempted pick from the midpoint, revealing a tilt. The demos then include lifting the bar from the correct CoM, resulting in a balanced lift. For the learner, success is counted if the bar is picked in a balanced way in the second try. We utilize 272 training demonstrations.

\subsection{Hardware}
We use operational space control~\cite{khatib1987unified} to relay torque commands to the robot.
$$\tau = J^T M_x(q) \left[ K_pe - K_d\dot{e} \right] + N^T \tau_0$$
Here, $e$ corresponds to pose error, $M_x$ is the task space inertia matrix, and $N$ is a nullspace projector. Our teleoperation data ($N=300$) is collected at 50Hz with 6D rotation, 3D position, and continuous gripper values. We predict 10D action targets in the same format as our observations. We perform Gram-Schmidt orthonormalization on the 6D rotation component of the predicted actions before they are fed into the operational space controller running at 1kHz. 

\begin{wraptable}{r}{0.51\textwidth}
\centering
\small
\caption{Success vs. Supervision}
\label{tab:supervision}
\begin{tabular}{lrrrr}
\toprule
Supervision Target & CD & DR & BB & Average\\
\midrule
Point & 63 & 100 & 88 & 84 \\
Patch & 92 & 100 & 89 & \textbf{94} \\
Image & 0 & 100 & 91 & 64 \\
\bottomrule
\end{tabular}
\end{wraptable}

\textbf{\half} We construct this task where the objective is to equally partition a set of $N \in \{2, 4\}$ 4cm x 4cm x 4cm cubes into two boxes that are too tall to see into with the global camera. As depicted in \Cref{fig:teleop}, the teleoperator follows a pattern for placing cubes into the bin which necessitates having memory to determine which box to put the next cube inside. We evaluate on the initial configuration shown (i.e. 2 cubes in the middle column). We show results in \Cref{fig:real-results} for evaluation across $20$ trials. We find that \dpnc{} gets around $50\%$ success since half the time it will put both in the right bin, or do the task correctly. This is since it does not have memory of whether it started with the 4 cube initial state or 2 cube initial state when it is placing the last cube. When we do endow the model with memory in \dphist{} and \keyframe, we find degradation of performance. The common failure mode was an inability to correctly pick and place the cubes by grasping in between the cubes or not grasping with the right depth. To ameliorate this, we trained these policies with wrist camera feeds observing some performance boost as reported in \Cref{fig:real-results}.  We use 75 demonstrations of the 4 configurations shown in \Cref{fig:teleop}, resulting in 300 trajectories. Wrist camera feeds were included by simply training a CNN encoder and concatenating it to the global camera encoding. For \keyframe{} and \wksp{}, we pruned erroneous detections of events at $t<30$ that sometimes appeared in training data. Furthermore, we simply used constant positions for the transient patches (initialized where the cubes start).

\section{Ablation Studies}
\label{app_ablation}

\textbf{Supervision Type} We also choose to ablate across different reconstruction targets to see which one forms the best latent representation for the downstream task. Instead of predicting the features for the patch which overlaps where MolmoPoint points, we can directly predict the 2D points during workspace model training. Another option is to predict the features across the whole image rather than the patch of interest.  We do this full-image supervision by predicting the average pooled patch feature. Our hypothesis is that point-based supervision is too lossy of a compression which removes relevant details such as rotations (which may be relevant for grasping) or visual features that could be relevant to the downstream task. Full-image supervision, on the other hand, can mix a lot of irrelevant information into the reconstruction target, which can cause potentially spurious correlations or simply increase the noise in our representations. These hypotheses are supported by our results in \Cref{tab:supervision} where we see that point-based supervision is strong yet underperforms the patch-based system. Meanwhile, the full-image system performs exceptionally well, but catastrophically breaks down in \cubedrop, presumably because this task requires the most fine-grained perception since cubes are spawned in randomized locations.

\textbf{VLM Detection} In \Cref{fig:generalization} we include the event firing plot of the VLM for \half{} that results from using the prompts specified for that task across every frame. It shows when VLM-prompted keyframes are detected and reveals an interesting pattern. First, there is a large amount of noisiness embedded in this detection process, which can be detrimental at runtime for test-time keyframe selection methods. For example, \citet{mark2026bpplongcontextrobotimitation} use ``rising-edges'' as timesteps to include keyframes, which would result in an inconsistent number of keyframes for every time. While this issue can be softened by higher quality VLM API calls, it requires careful prompt tuning and tuned VLM polling intervals. However, by having access to the full trajectory during train-time, offline saliency-driven labeling enables a more noise-resistant pipeline that is robust to lower quality VLM detection labels and more consistent in timings (since it's a median). The PCA plot also depicts a smooth representation across time, giving a gradually changing representation across the expert demonstration. In \Cref{fig:decoder_firing}, we show the slot probabilities for the trained workspace model on \half{} as well as the frames at which the slot probabilities cross a threshold. It can be observed that the \wksp{} latent event firing corresponds to highly semantically similar frames (i.e., the gripper dropping a cube in the box) which displays consistency. We hypothesize this consistency is derived from the smoothing that can occur from training on many examples. So, not only does a median-based event time labeling system give resistant labels, but the distillation process also creates a smoothing effect.

\end{document}